\pdfoutput=1

\documentclass[11pt]{article}

\usepackage[final]{acl}

\usepackage{times}
\usepackage{latexsym}

\usepackage[T1]{fontenc}

\usepackage[utf8]{inputenc}

\usepackage{microtype}

\usepackage{inconsolata}

\usepackage{graphicx}
\usepackage{hyperref}       
\usepackage{url}            
\usepackage{booktabs}       
\usepackage{amsfonts}       
\usepackage{nicefrac}       
\usepackage{microtype}      
\usepackage{xcolor}         
\usepackage{multirow}
\usepackage{float}
\usepackage{amsmath}
\usepackage{subcaption}
\usepackage{wrapfig}
\usepackage[most]{tcolorbox}
\usepackage{enumitem}
\usepackage{stfloats}
\usepackage{tcolorbox}   
\tcbuselibrary{skins}    

\usepackage{colortbl}
\usepackage{xcolor}
\definecolor{mygreen1}{rgb}{0.9686, 0.9804, 0.9843}  
\definecolor{mygreen2}{rgb}{0.7961, 0.8745, 0.9216}  
\definecolor{mygreen3}{rgb}{0.6235, 0.7647, 0.8588}  
\definecolor{mygreen4}{rgb}{0.3686, 0.5961, 0.7608} 

\usepackage{mathtools}
\usepackage{amsthm}
\theoremstyle{plain}
\newtheorem{findings}{Findings}
\newtheorem{type}{Type}

\newtheorem{desideratum}{Desideratum}
    
\title{\textit{Cut Your Losses!} Learning to Prune Paths Early for Efficient Parallel Reasoning}

\author{
   Jiaxi Bi$^{1,3}$\thanks{~~Equal contribution; alphabetical by last name.}\thanks{~~Work done during interning at CUHK-Shenzhen.}
   \quad  Tongxu Luo$^{1,2*}$  
   \quad  Wenyu Du$^{4}$ 
   \quad  \textbf{Zhengyang Tang$^{1}$} 
   \quad \textbf{Benyou Wang$^{1,2}$\thanks{~~Corresponding author.}} \\
  $^{1}$The Chinese University of Hong Kong, Shenzhen \\
  $^{2}$Shenzhen Loop Area Institute \\
  $^{3}$USTB \quad
  $^{4}$DualityRL
  \\
  \texttt{jiaxibi@xs.ustb.edu.cn \quad tongxuluo@cuhk.edu.cn \quad wangbenyou@cuhk.edu.cn} 
}

\begin{document}
\maketitle
\begin{abstract}

Parallel reasoning enhances Large Reasoning Models (LRMs) but incurs prohibitive costs due to futile paths caused by early errors. 
To mitigate this, path pruning at the prefix level is essential, yet existing research remains fragmented without a standardized framework. 
In this work, we propose the first systematic taxonomy of path pruning, categorizing methods by their signal \textit{source} (internal vs. external) and \textit{learnability} (learnable vs. non-learnable). 
This classification reveals the unexplored potential of \textit{learnable internal methods}, motivating our proposal of \textbf{STOP} (\textbf{S}uper \textbf{TO}ken for \textbf{P}runing). 
Extensive evaluations across LRMs ranging from 1.5B to 20B parameters demonstrate that STOP achieves superior effectiveness and efficiency compared to existing baselines. 
Furthermore, we rigorously validate the scalability of STOP under varying compute budgets—for instance, boosting GPT-OSS-20B accuracy on AIME25 from 84\% to nearly 90\% under fixed compute budgets. 
Finally, we distill our findings into formalized empirical guidelines to facilitate optimal real-world deployment. 
Code, data and models are available at \url{https://bijiaxihh.github.io/STOP}.
\end{abstract}

\section{Introduction}\label{sec:Introduction}
Parallel reasoning has established itself as a standard paradigm for solving complex problems~\cite{openai2024a,wang2025survey_parallel_reasoning}.
The core principle is to sample multiple independent reasoning paths and subsequently aggregate them to derive a robust consensus.
However, this accuracy gain comes at a prohibitive cost.
Generating dozens or even hundreds of trajectories per query increases computational overhead by orders of magnitude~\cite{jin2025energy} and escalates inference costs to nearly \$6 per query~\cite{nvidia2025inferencecost}.


\begin{figure}[t]
    \centering
    \vspace{-2mm}
    \includegraphics[width=\linewidth]{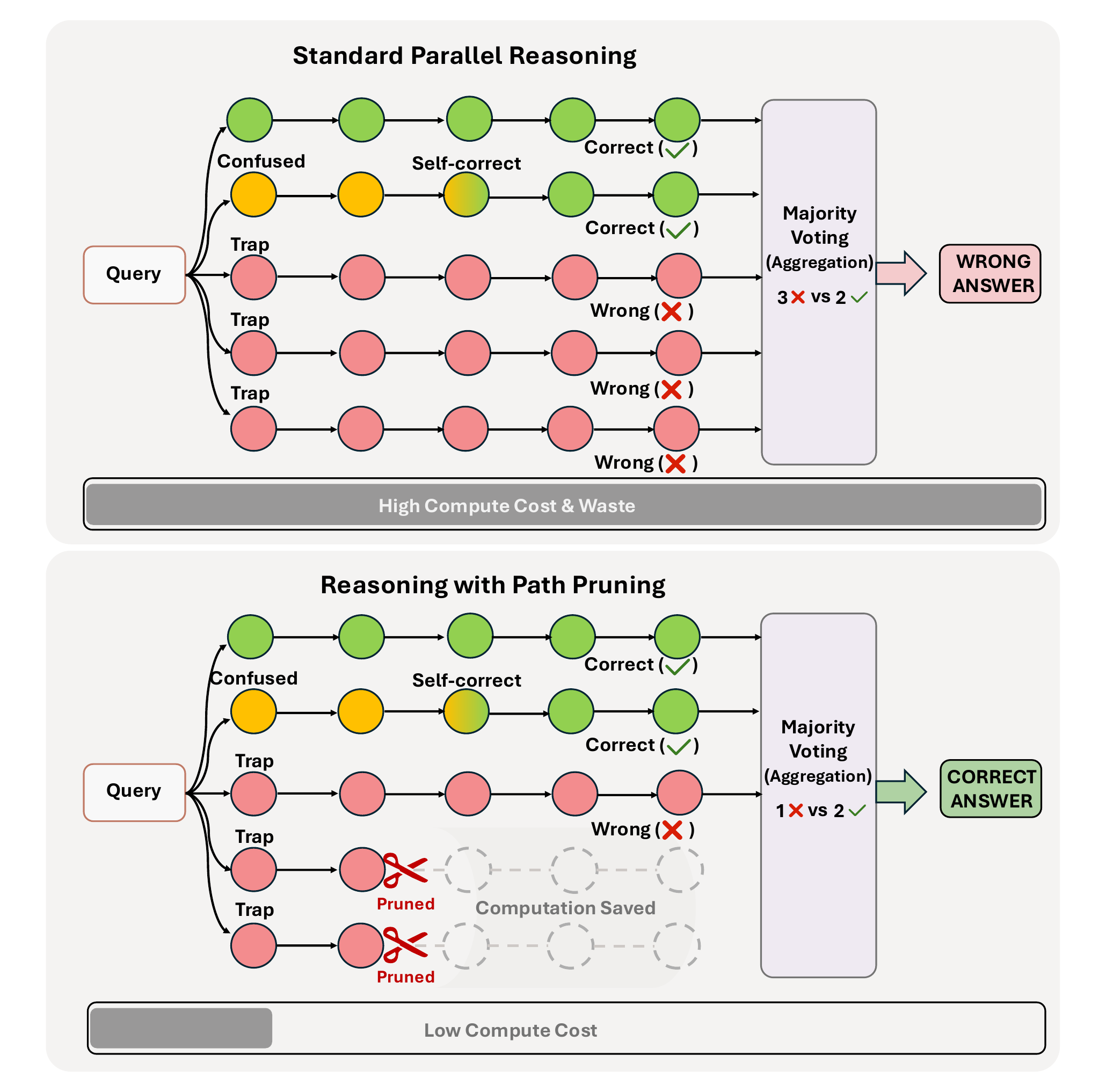} 
    \caption{The necessity of pruning early. Early errors often lead to irreversible failure. Pruning these futile paths early not only saves computation but also purifies the candidate set for better consensus.}
    \label{fig:motivation}
    \vspace{-6mm}
\end{figure}

\paragraph{Why Prune Early in Parallel Reasoning?}
Crucially, recent studies~\cite{luo2025peers,hassid2025short} reveal that this extensive computation is largely squandered: \textbf{not every path contributes to the solution}.
Many trajectories are flawed from inception, yet they consume equal resources to generate and subsequently pollute the final answer aggregation.
As illustrated in Figure~\ref{fig:motivation}, once a reasoning path begins with a flawed prefix, the LRM struggles to self-correct, inevitably spiraling into a futile trajectory~\cite{luo2025peers}.
Consequently, identifying and terminating these unpromising paths at the \textit{prefix level}—a technique known as \textbf{path pruning~(or prefix rejection)}—is essential.

\begin{figure}[ht]
  \centering
  \includegraphics[width=\linewidth]{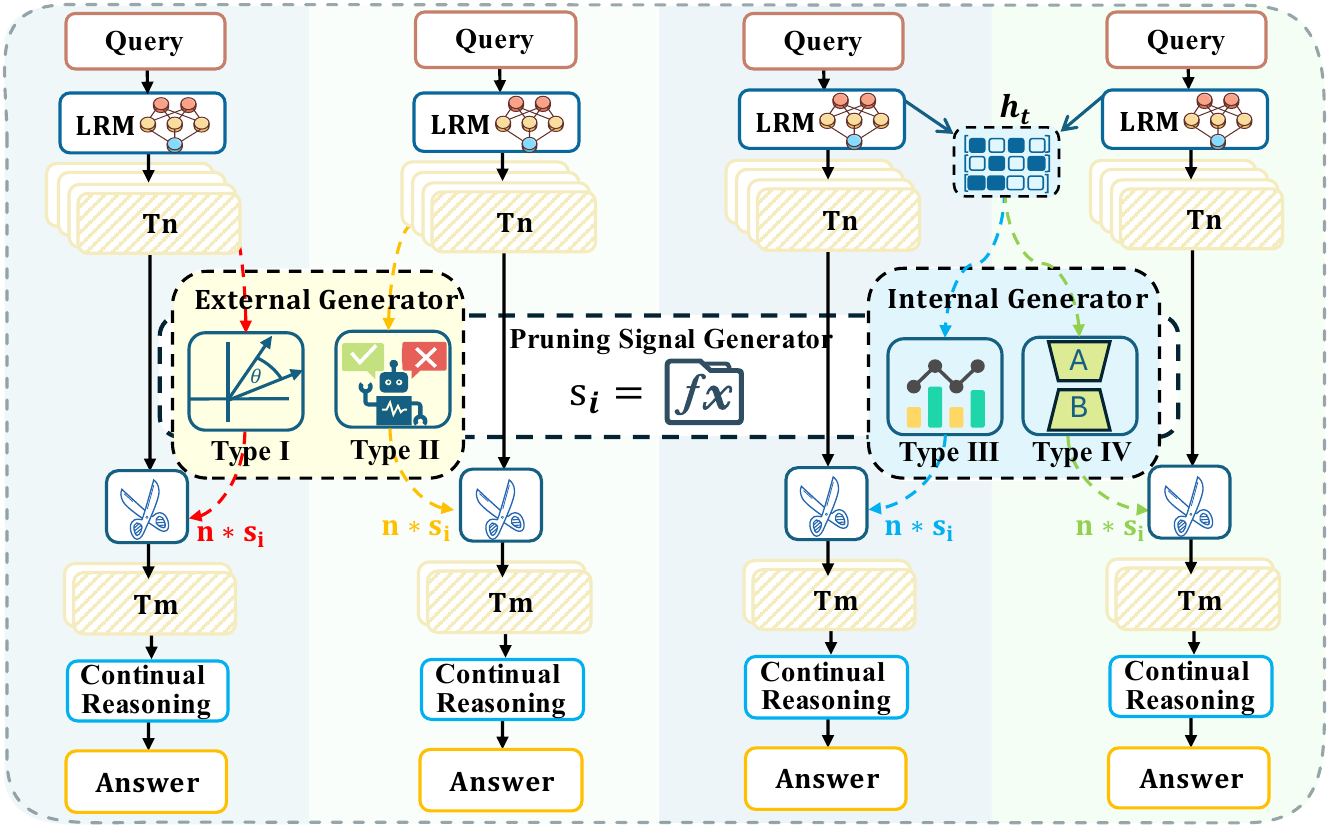}
  \caption{The proposed taxonomy of path pruning.}
  \label{fig:main}
\vspace{-4mm}
\end{figure}

\paragraph{A Unified Taxonomy}
While existing methods attempt to filter paths using auxiliary reward models~\cite{liao2025lost}, internal confidence~\cite{fu2025deepthink}, or semantic redundancy~\cite{hong2025slimsc}, they lack a standardized evaluation protocol, leading to fragmented research.
So first, we propose the first systematic taxonomy of path pruning, classifying methods based on the \textit{source} (internal vs. external) and \textit{learnability} (learnable vs. non-learnable) of their signals (see Figure~\ref{fig:main}).
This taxonomy reveals a significant research gap: the unexplored potential of \textit{learnable internal methods}.
Conceptually, learnable internal methods offer unique advantages, as learning enables task-specific accuracy gains, while internal signals provide early, fine-grained indicators of reasoning failure without incurring extra computational overhead.
To bridge this gap, we introduce \textbf{STOP} (\textbf{S}uper \textbf{TO}ken for \textbf{P}runing), the first efficient instantiation of this paradigm.
Extensive evaluations demonstrate that STOP outperforms existing baselines in both effectiveness and efficiency.

\paragraph{Further Evaluation and Empirical Analysis}
Despite the promise of path pruning, its widespread adoption is currently hindered by unverified scalability across varying computational budgets and model sizes; and the absence of empirical guidelines for determining optimal pruning configurations in real-world scenarios.
To overcome them, we rigorously validate the utility of path pruning in practical settings.
We conduct extensive experiments across diverse model sizes (1.5B to 20B) and compute budgets, confirming that STOP exhibits robust scalability.
Moreover, we distill our empirical analysis into actionable guidelines, providing a formalized method to determine the optimal retention ratio for varying resource constraints.

\paragraph{Contributions}
In summary, this work makes four primary contributions:
(1) We present the first systematic investigation and taxonomy of path pruning.
(2) We propose STOP, a novel pruning method based on learnable internal signals.
(3) We provide a comprehensive evaluation demonstrating STOP's superior scalability and effectiveness.
(4) We establish empirical guidelines to support the practical implementation of path pruning.

\section{A Unified Taxonomy of Path Pruning}\label{sec:Four Types}
\subsection{Problem Definition}
\label{subsec:definition}

Consider a LRM $\Theta$ and an input query $x$, parallel reasoning improves accuracy by generating $N$ independent trajectories $T = \{\tau_i\}_{i=1}^N$, where $\tau_i \sim P_{\Theta}(x)$, and aggregating them through a consensus strategy, such as majority voting.
The final prediction $\hat{y}$ is typically computed as:
\begin{equation}\small
    \hat{y} = \text{vote}(\{\tau_i\}_{i=1}^N).
\end{equation}
However, generating $N$ complete trajectories incurs a linear computational cost ($C \propto N$).
To mitigate this cost, path pruning aims to identify and discard unpromising trajectories early in the decoding process.

\paragraph{The Path Pruning Formulation}
Formally, we define a checkpoint at length $L_\text{prefix}$ where the generation is paused.
At this stage, the model has produced a set of prefixes $\mathcal{P} = \{p_i\}_{i=1}^N$.
The core of path pruning is a \textbf{pruning signal generator} $S$, which maps each prefix to a scalar score representing its potential correctness:
\begin{equation}\small
    s_i = S(p_i \mid x, \Theta),
\end{equation}
where $s_i \in [0, 1]$ denotes the pruning signal.
Based on these signals, we retain only the top-$k$ promising paths (where $k \ll N$) for full completion, discarding the rest.
The final aggregated answer is then derived exclusively from this pruned subset:
\begin{equation}
    \hat{y}_\text{pruned} = \text{vote}(\{\text{finish}(p_i) \mid s_i \in \{s_j\}_{j=1}^k\}).
\end{equation}
So, the objective of path pruning is to design an $S$ that maximizes $\hat{y}_\text{pruned}$'s accuracy while minimizing the computational cost (the number of generated tokens).
Therefore, the design of $S$ dictates the effectiveness of the entire framework.

\subsection{A Unified Taxonomy of Pruning Signal Generators}
\label{sec:taxonomy}

\begin{table*}[t]
\vspace{-12mm}
\centering
\caption{A Unified Taxonomy of Path Pruning Methods. We categorize methods based on the pruning signal source and learnability. \textbf{Type~\ref{type:module}} satisfies both \textbf{Desideratum~\ref{desideratum:internal}} (Internal) and \textbf{Desideratum~\ref{desideratum:learnable}} (Learnable).}
\label{tab:taxonomy}
\renewcommand{\arraystretch}{1.5} 
\resizebox{0.9\linewidth}{!}{
\begin{tabular}{l|c|c}
\toprule
 & \textbf{Non-Learnable} & \textbf{Learnable} \\
 &  & (Desideratum~\ref{desideratum:learnable}) \\ \midrule

\textbf{External Source} & \textbf{Type~\ref{type:heuristic}} & \textbf{Type~\ref{type:aux}} \\
 & \begin{tabular}[c]{@{}c@{}}
\textbf{SlimSC}~\cite{hong2025slimsc}
\end{tabular} & \begin{tabular}[c]{@{}c@{}}
\textbf{DeepPrune}~\cite{tu2025deepprune}, \textbf{LaBoR}~\cite{liao2025lost} \\
\textbf{ThinkPRM}~\cite{khalifa2025process}, \textbf{MAV}~\cite{lifshitz2025multi}
\end{tabular} \\ \midrule

\textbf{Internal Source} & \textbf{Type~\ref{type:stats}} & \textbf{Type~\ref{type:module}} \\
(Desideratum~\ref{desideratum:internal}) & \begin{tabular}[c]{@{}c@{}}
\textbf{DeepConf}~\cite{fu2025deepthink}, 
\textbf{AdaDec}~\cite{he2025adadec} \\
\textbf{Think Just Enough}~\cite{sharma2025thinkjust}
\end{tabular} & \begin{tabular}[c]{@{}c@{}}
\textbf{STOP (Ours)},
\textbf{OTV}\footnotemark~\cite{zhuang2026one}
\end{tabular} \\ \bottomrule
\end{tabular}
}
\vspace{-4mm}
\end{table*}

As defined in Section~\ref{subsec:definition}, the efficacy of path pruning hinges entirely on the quality of the pruning signal generator $S$.
While the function of $S$ is consistent—scoring prefixes—existing methods differ fundamentally in \textit{how} this signal is produced.
To systematically evaluate these approaches, we categorize them based on two critical dimensions: the \textit{source} of the signal (External vs. Internal) and the \textit{learnability} of the generator (Learnable vs. Non-learnable), as summarized in Table~\ref{tab:taxonomy}.

\paragraph{Two Desiderata for Signal Generators}
Before categorizing specific methods, we establish two desiderata for an ideal signal generator:

\begin{desideratum}\label{desideratum:internal}
\textbf{Internal Source} An ideal $S$ should leverage the rich, high-dimensional internal states of the LRM.
\end{desideratum}
Internal signals contain fine-grained information about uncertainty and reasoning dynamics that are often lost in the final text output used by external methods.

\begin{desideratum}\label{desideratum:learnable}
\textbf{Learnability} An ideal $S$ should be trainable to adapt to specific data distributions.
\end{desideratum}
Learnable parameters allow the generator to capture complex, non-linear patterns of error that rigid, pre-defined heuristics cannot model.

Based on these axes, we classify existing works into four distinct types.

\paragraph{External Signal Source}
Methods in this category derive pruning signals from the generated textual output or by querying separate models. They fail to satisfy Desideratum~\ref{desideratum:internal}.

\begin{type}\label{type:heuristic} \textbf{Surface Heuristics}
These methods rely on human-designed rules (e.g. similarity) applied to the surface form of the generated text.
\end{type}
While computationally cheap, these heuristics are rigid and blind to the model's actual confidence.
To overcome these, the next type introduces learnability into the external evaluation process.

\begin{type}\label{type:aux} \textbf{External Judges}
These approaches employ a separate, trained model to evaluate the reasoning path.
\end{type}
Although they satisfy Desideratum~\ref{desideratum:learnable}, they incur significant computational overhead due to the need for additional model inference and fail to access the LRM's internal certainty.
To overcome this rigidity, the next category introduces learnability into the external evaluation process.

\paragraph{Internal Signal Source}
Methods in this category extract signals directly from the LRM's internal states, accessing to richer information (satisfying Desideratum~\ref{desideratum:internal}).

\begin{type}\label{type:stats} \textbf{Raw Confidence}
This paradigm utilizes intrinsic metrics directly derived from the decoding process, such as perplexity or token probability.
\end{type}
However, these methods rely on fixed definitions of confidence, violating Desideratum~\ref{desideratum:learnable}; raw probability does not always correlate with reasoning correctness.

\begin{type}\label{type:module} \textbf{Learned Intuition}
The final category represents the intersection of both desiderata: a trainable module inserted into the LRM to process internal states.
\end{type}
This approach can leverage rich hidden representations (Internal) while adapting to the specific error patterns of the task (Learnable).

\footnotetext{We respectfully note that OTV is our concurrent work, and they are also the first Type IV method.}


\section{Methodology: Super Token for Pruning}


As established in our taxonomy, Type~\ref{type:module} represents the ideal pruning paradigm but remains unexplored.
In this section, we introduce \textbf{STOP} (\textbf{S}uper \textbf{TO}ken for \textbf{P}runing), the first efficient instantiation of this paradigm.
We delineate the motivation in Section~\ref{subsec:motivation}, followed by the architectural design and workflow in Section~\ref{subsec:architecture}.

\subsection{Motivation for Type~\ref{type:module} Pruning}\label{subsec:motivation}
As illustrated in Figure~\ref{fig:main}, prior methods compromise on either information richness or adaptability.
Type~\ref{type:aux} suffers from high latency, while Type~\ref{type:stats} lacks the capacity to model complex error patterns.
Type~\ref{type:module} represents an ideal optimum: it combines the \textit{efficiency} of accessing internal states with the \textit{adaptability} of learnable parameters.
However, this type remains unexplored due to the challenge of designing a module that extracts these signals without disrupting the LRM's generative capabilities.

\begin{figure*}[ht]
\vspace{-12mm}
    \centering
    \includegraphics[width=\textwidth]{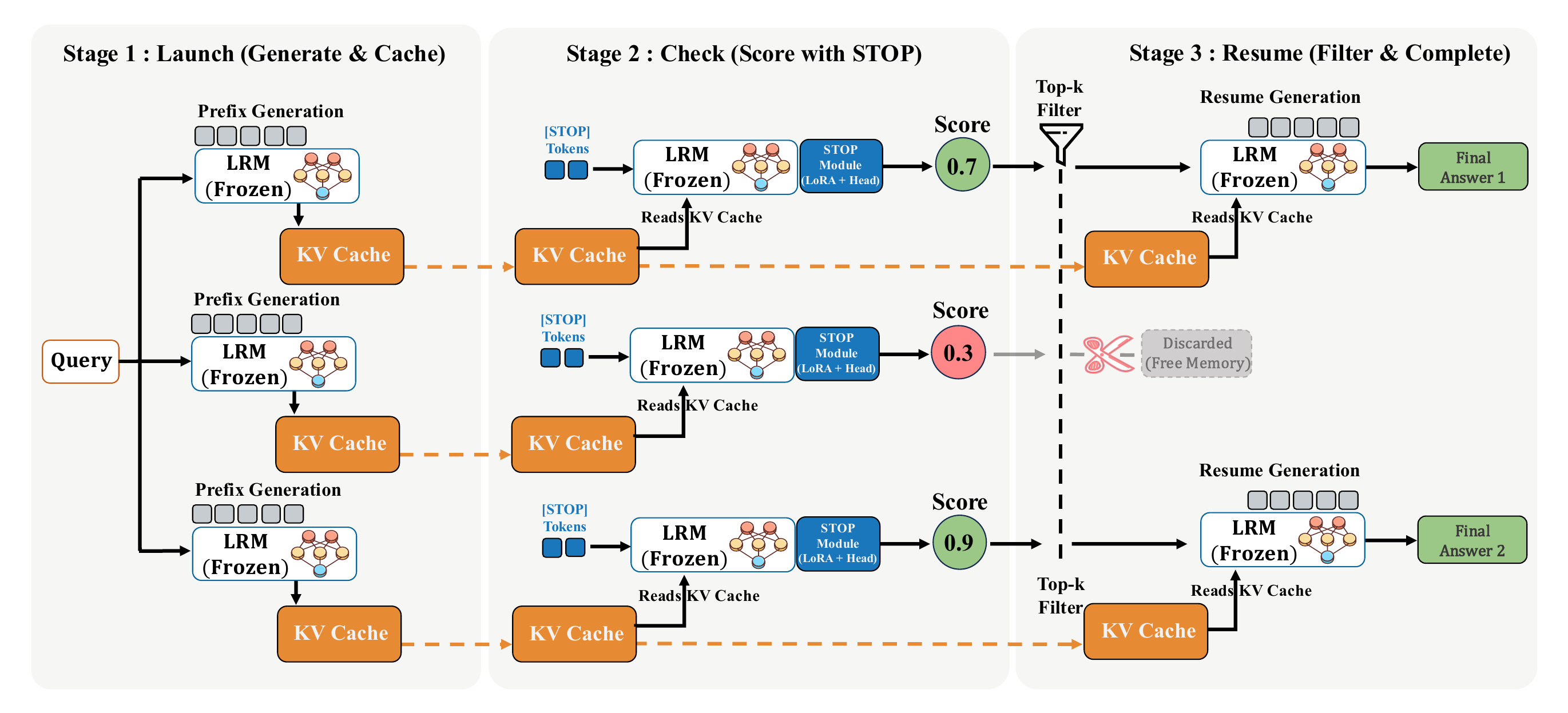}
    \caption{The inference process comprises three stages: caching initial prefixes (\textbf{Launch}), scoring them via the \textbf{STOP} module (\textbf{Check}), and completing only the top-ranked candidates (\textbf{Resume}).}
    \label{fig:method_STOP}
\vspace{-4mm}
\end{figure*}

\subsection{Instantiation of Type~\ref{type:module} Pruning: STOP}
\label{subsec:architecture}

To instantiate this type, we design \textbf{STOP} as a lightweight, non-invasive module that integrates seamlessly with the backbone LRM.

\paragraph{Components}
We augment the fixed LRM $\Theta$ with three learnable components:
(1) \textbf{A Super Token (\texttt{[STOP]})} added to the vocabulary, acting as a specialized query vector to aggregate information;
(2) \textbf{A Critique Adapter} LoRA ($\theta_\text{LoRA}$), activated only when processing the \texttt{[STOP]} token to extract error-specific features without altering the LRM's general reasoning capabilities;
(3) \textbf{A Classification Head} ($W_\text{cls}$), which projects the hidden state of the \texttt{[STOP]} token to a scalar probability.

This design ensures \textbf{modularity}: the original parameters $\Theta$ remain frozen, preserving the LRM's generative capability while enabling efficient parameter-efficient fine-tuning (PEFT).

\paragraph{Training: Learn to Use Internal Information}
The goal of training is simple: teach the model to distinguish promising prefixes from futile ones.
Formally, for a prefix $p_i$, we derive a soft label $s^{mc}_i \in [0,1]$ via Monte Carlo estimation (details in Appendix~\ref{app:data_construction}).
The training process involves two steps:
First, we compute the KV cache of the prefix using the frozen LRM: $\mathcal{C}_{p_i} = \text{LRM}(p_i; \Theta)$.
Second, we append a sequence of learnable \texttt{[STOP]} tokens, denoted as $T_s$, and process them using the LoRA-augmented model.
The final hidden state $h_i$ is fed into the classifier to minimize the soft binary cross-entropy loss:
\begin{equation}\small
\begin{array}{ll}
    \mathcal{L} = & - [ s_i^{mc} \log \sigma (W_{cls}h_i) \\
    & +  (1-s_i^{mc}) \log (1-\sigma (W_{cls}h_i)) ],
\end{array}
\end{equation}
where $h_i = \text{LRM}(T_s \mid \mathcal{C}_{p_i}; \Theta, \theta_{\text{LoRA}})_{-1}$.

\paragraph{Training Cost}
Constructing the MC supervision requires sampling multiple continuations per prefix to estimate $s^{mc}_i$ (e.g., $K=32$), which introduces an upfront computational cost during data construction. However, this cost is incurred only once, and the resulting STOP module is lightweight and reusable across tasks. To facilitate transparency and reproducibility, we provide detailed cost statistics in Appendix~\ref{app:training_cost} and will release the constructed dataset and trained checkpoints, allowing practitioners to bypass this step entirely. Importantly, this one-time cost is amortized during deployment, where STOP improves efficiency by pruning unpromising paths early.

\paragraph{Inference: ``Launch-Check-Resume''}
To efficiently prune paths without slowing down generation, we design a three-stage pipeline (Figure~\ref{fig:method_STOP}):

\textbf{Stage 1: Launch}
Instead of generating the full trajectories immediately, we first generate $N$ short prefixes (e.g., first 1024 tokens) for the query.
Crucially, we cache the internal states (KV Cache) of these prefixes.

\textbf{Stage 2: Check}
We append the \texttt{[STOP]} tokens to the cached prefixes.
The trained module reads the KV cache and outputs a quality score for each prefix.
\textit{Note:} This step is extremely fast because it processes only a few tokens (the \texttt{[STOP]} sequence) and reuses the heavy computation already done in Stage 1.

\textbf{Stage 3: Resume}
We rank the prefixes by their scores and apply a \textbf{Top-$k$ Filter}.
Futile paths are discarded immediately to free up memory.
Only the top-$k$ most promising prefixes are resumed and generated to completion to obtain the final answers.

\begin{table*}[ht!]
\vspace{-12mm}
\centering
\caption{
    Results of avg@k (avg@m|k) across various models and benchmarks. 
    The best result in each row is \textbf{bolded} and the second best is \underline{underlined}.
}
\label{tab:main_results}
\resizebox{\textwidth}{!}{
\begin{tabular}{llcc cc cc cc cc}
\toprule

\multirow{2}{*}{\textbf{Model}} & \multirow{2}{*}{\textbf{Dataset}} 
& \multicolumn{2}{c}{\textbf{No pruning (Baseline)}} 
& \multicolumn{2}{c}{\textbf{Type~\ref{type:heuristic} }} 
& \multicolumn{2}{c}{\textbf{Type~\ref{type:aux}}} 
& \multicolumn{2}{c}{\textbf{Type~\ref{type:stats}}} 
& \multicolumn{2}{c}{\textbf{Type~\ref{type:module}}} \\
\cmidrule(lr){3-4} \cmidrule(lr){5-6} \cmidrule(lr){7-8} \cmidrule(lr){9-10}\cmidrule(lr){11-12}
& & avg@64 ($\uparrow$) & Tokens ($\downarrow$)  & avg@8|64 ($\uparrow$) & Tokens (\% $\downarrow$) & avg@8|64 ($\uparrow$) & Tokens (\% $\downarrow$) & avg@8|64 ($\uparrow$) & Tokens (\% $\downarrow$)& avg@8|64 ($\uparrow$) & Tokens (\% $\downarrow$) \\

\midrule
\multirow{5}{*}{DS-Qwen-2.5-1.5B} 
& AIME24    & \cellcolor{mygreen1}30.10 & 782.3k  & \cellcolor{mygreen1}26.25 & 218.3k~(-72.09\%)  & \cellcolor{mygreen2}32.50 & 325.9k~(-58.34\%) & \cellcolor{mygreen2}\underline{32.92} & \underline{210.6k}~(-73.08\%)    & \cellcolor{mygreen3}\textbf{37.92} & \textbf{204.3k}~(-73.88\%)\\
& AIME25    & \cellcolor{mygreen1}22.76 & 784.8k  & \cellcolor{mygreen2}\underline{24.17} & 214.7k~(-72.64\%)  & \cellcolor{mygreen2}\underline{24.17} & 325.0k~(-58.59\%) & \cellcolor{mygreen2}23.75 & \underline{208.7k}~(-73.40\%)    & \cellcolor{mygreen3}\textbf{26.67} & \textbf{206.6k}~(-73.68\%)\\
& BRUMO25   & \cellcolor{mygreen2}30.99 & 774.6k  & \cellcolor{mygreen1}29.58 & 212.8k~(-72.53\%)  & \cellcolor{mygreen2}\underline{31.67} & 325.6k~(-57.96\%) & \cellcolor{mygreen2}31.67 & \underline{209.7k}~(-72.93\%)    & \cellcolor{mygreen3}\textbf{33.75} & \textbf{204.4k}~(-73.61\%) \\
& HMMT25    & \cellcolor{mygreen1}15.05 & 856.4k  & \cellcolor{mygreen1}15.83 & 224.2k~(-73.82\%)  & \cellcolor{mygreen1}15.00 & 337.2k~(-60.63\%) & \cellcolor{mygreen2}\underline{17.08} & \underline{220.9k}~(-74.21\%)    & \cellcolor{mygreen3}\textbf{17.92} & \textbf{215.5k}~(-74.84\%)\\
& GPQA-D    & \cellcolor{mygreen2}33.08 & 550.9k  & \cellcolor{mygreen1}32.32 & 187.1k~(-66.03\%)  & - & - & \cellcolor{mygreen2}\underline{34.98} & \underline{180.4k}~(-67.25\%)    & \cellcolor{mygreen3}\textbf{48.42} & \textbf{179.4k}~(-67.43\%) \\

\midrule
\multirow{5}{*}{DS-Qwen-2.5-7B} 
& AIME24    & \cellcolor{mygreen2}54.69 & 666.2k  & \cellcolor{mygreen1}53.75 & 202.7k~(-69.58\%)  & \cellcolor{mygreen2}54.58 & 312.5k~(-53.09\%) & \cellcolor{mygreen2}\underline{55.00} & \underline{197.4k}~(-70.38\%)    & \cellcolor{mygreen3}\textbf{61.67} & \textbf{189.0k}~(-71.63\%) \\
& AIME25    & \cellcolor{mygreen2}39.67 & 703.0k  & \cellcolor{mygreen1}35.42 & 207.4k~(-70.50\%)  & \cellcolor{mygreen2}39.17 & 317.6k~(-54.82\%) & \cellcolor{mygreen2}\underline{41.67} & \underline{202.6k}~(-71.18\%)    & \cellcolor{mygreen3}\textbf{42.50} & \textbf{197.5k}~(-71.91\%) \\
& BRUMO25   & \cellcolor{mygreen2}50.99 & 656.6k  & \cellcolor{mygreen1}50.00 & 199.3k~(-69.64\%)  & \cellcolor{mygreen2}51.25 & 312.1k~(-52.46\%) & \cellcolor{mygreen2}\underline{52.92} & \underline{194.5k}~(-70.38\%)    & \cellcolor{mygreen3}\textbf{56.67} & \textbf{190.2k}~(-71.03\%)\\
& HMMT25    & \cellcolor{mygreen1}23.91 & 808.9k  & \cellcolor{mygreen2}\underline{25.00} & 219.5k~(-72.87\%)  & \cellcolor{mygreen1}23.33 & 330.8k~(-59.11\%) & \cellcolor{mygreen2}24.58 & \underline{216.1k}~(-73.28\%)    & \cellcolor{mygreen3}\textbf{27.08} & \textbf{211.6k}~(-73.84\%)\\
& GPQA-D    & \cellcolor{mygreen1}45.95 & 443.8k  & \cellcolor{mygreen2}\underline{47.09} & 173.6k~(-60.89\%)  & - & - & \cellcolor{mygreen2}46.02 & \underline{166.7k}~(-62.43\%)    & \cellcolor{mygreen3}\textbf{55.75} & \textbf{165.9k}~(-62.61\%)\\

\midrule
\multirow{5}{*}{DS-Qwen-3-8B} 
& AIME24    & \cellcolor{mygreen1}76.93 & 1361k   & \cellcolor{mygreen2}77.50 & 290.4k~(-78.67\%)  & \cellcolor{mygreen2}\underline{78.75} & 398.4k~(-70.73\%) & \cellcolor{mygreen2}78.33 & \underline{284.3k}~(-79.11\%)    & \cellcolor{mygreen3}\textbf{79.17} & \textbf{279.0k}~(-79.51\%)\\
& AIME25    & \cellcolor{mygreen2}70.68 & 1427k   & \cellcolor{mygreen1}69.17 & 297.2k~(-79.18\%)  & \cellcolor{mygreen2}\underline{72.50} & 408.4k~(-71.39\%) & \cellcolor{mygreen2}70.42 & \underline{291.2k}~(-79.60\%)    & \cellcolor{mygreen3}\textbf{72.92} & \textbf{290.9k}~(-79.62\%)\\
& BRUMO25   & \cellcolor{mygreen2}75.00 & 1320k   & \cellcolor{mygreen2}\underline{76.25} & 284.8k~(-78.43\%)  & \cellcolor{mygreen2}75.83 & 394.9k~(-70.10\%) & \cellcolor{mygreen1}74.58 & \underline{280.2k}~(-78.78\%)    & \cellcolor{mygreen3}\textbf{78.75} & \textbf{277.5k}~(-78.98\%)\\
& HMMT25    & \cellcolor{mygreen2}51.04 & 1601k   & \cellcolor{mygreen1}50.00 & 322.1k~(-79.88\%)  & \cellcolor{mygreen1}50.83 & 427.8k~(-73.28\%) & \cellcolor{mygreen2}\underline{51.25} & \underline{314.0k}~(-80.38\%)    & \cellcolor{mygreen3}\textbf{54.58} & \textbf{311.7k}~(-80.53\%)\\
& GPQA-D    & \cellcolor{mygreen1}56.87 & 652.6k  & \cellcolor{mygreen2}57.07 & 201.0k~(-69.20\%)  & - & - & \cellcolor{mygreen2}\underline{58.78} & \underline{196.6k}~(-69.86\%)    & \cellcolor{mygreen3}\textbf{63.32} & \textbf{193.5k}~(-70.35\%)\\

\midrule
\multirow{5}{*}{GPT-OSS-20B}
& AIME24    & \cellcolor{mygreen2}75.26 & 594.2k  & \cellcolor{mygreen1}73.33 & 196.6k~(-66.91\%)  & \cellcolor{mygreen2}\underline{76.25} & 299.8k~(-49.55\%) & \cellcolor{mygreen1}75.00 & \underline{187.0k}~(-68.54\%)    & \cellcolor{mygreen3}\textbf{77.50} & \textbf{184.4k}~(-68.98\%) \\
& AIME25    & \cellcolor{mygreen2}70.99 & 673.4k  & \cellcolor{mygreen1}69.17 & 205.1k~(-69.54\%)  & \cellcolor{mygreen1}69.17 & 311.7k~(-53.71\%) & \cellcolor{mygreen2}\underline{70.83} & \underline{197.7k}~(-70.64\%)    & \cellcolor{mygreen3}\textbf{75.42} & \textbf{191.1k}~(-71.62\%)\\
& BRUMO25   & \cellcolor{mygreen2}68.02 & 575.6k  & \cellcolor{mygreen2}\underline{69.17} & 187.6k~(-67.41\%)  & \cellcolor{mygreen1}66.25 & 298.8k~(-48.09\%) & \cellcolor{mygreen1}67.92 & \textbf{179.0k}~(-68.90\%)       & \cellcolor{mygreen3}\textbf{70.00} & \underline{183.6k}~(-68.11\%) \\
& HMMT25    & \cellcolor{mygreen2}\underline{48.13} & 910.8k & \cellcolor{mygreen1}45.83 & 236.3k~(-74.06\%)  & \cellcolor{mygreen1}45.42 & 336.9k~(-63.01\%) & \cellcolor{mygreen1}46.25 & \underline{228.0k}~(-74.97\%)    & \cellcolor{mygreen3}\textbf{52.92} & \textbf{216.1k}~(-76.27\%)\\
& GPQA-D    & \cellcolor{mygreen1}65.55 & 277.2k  & \cellcolor{mygreen2}66.41 & 151.8k~(-45.24\%)  & - & - & \cellcolor{mygreen2}\underline{68.43} & \underline{145.9k}~(-47.36\%)    & \cellcolor{mygreen3}\textbf{77.46} & \underline{143.4k}~(-48.26\%)\\

\bottomrule
\end{tabular}
}
\vspace{-4mm}
\end{table*}

\section{A Close Look at Path Pruning through the Lens of Signal Generators}
\label{sec:Comprehensive}


\subsection{On the Effectiveness of Pruning}
\label{subsec:On the Effectiveness of Pruning}
To systematically evaluate the effectiveness of four types of pruning signal generators in our taxonomy, we conduct extensive experiments on five reasoning benchmarks.
We employ a diverse suite of LRMs ranging from 1.5B to 20B parameters, specifically the DeepSeek-R1-Distill-Qwen series~\cite{guo2025deepseek} and gpt-oss-20b~\cite{openai2025gptoss}.

\paragraph{Standardized protocol.}
To ensure a fair comparison, we establish a standardized evaluation protocol: for each query, we generate $64$ initial reasoning paths.
We prune these to the top $8$ candidates.
For each $S$, we apply pruning at \textbf{2,048 tokens} to rigorously evaluate their ability to identify futile paths with limited context.

\paragraph{Evaluation metrics.}
We report two metrics: \textbf{(1) avg@k}, defined as the average accuracy over the $k$ paths. 
In the context of pruning, we denote this metric as \textbf{avg@m|k} (selecting $m$ from $k$). 
Since random pruning theoretically yields an average accuracy equivalent to the no-pruning baseline, \textbf{a pruning method is considered effective only if its avg@m|k surpasses the baseline avg@k}, thereby indicating a higher density of correct answers in the selected subset.
\textbf{(2) total tokens}, which is used to quantify computational cost.
We calculate the relative token reduction $\Delta$ as:
\begin{equation}\small
    \Delta = \frac{\text{Tokens}_{\text{original}} - \text{Tokens}_{\text{pruned}}}{\text{Tokens}_{\text{original}}} \times 100\%.
\end{equation}
We list the detailed experimental settings, including infrastructure and hyperparameters in Appendix~\ref{app:exp_setup}.

\paragraph{Performance Hierarchy across Four Types Pruning}
As presented in Table~\ref{tab:main_results}, while most pruning signals demonstrate effectiveness, we observe distinct performance hierarchies. 
First, internal-based generators (Type~\ref{type:stats} and Type~\ref{type:module}) consistently outperform external-based ones (Type~\ref{type:heuristic} and Type~\ref{type:aux}). 
This advantage stems from their access to internal LRM states—such as hidden states and KV caches—which encode significantly richer representations than the constrained natural language outputs used by external methods. 
Second, learnable generators (Type~\ref{type:module} and Type~\ref{type:aux}) surpass non-learnable baselines, as both leverage training data to detect reasoning errors at early stages; we further validate this by explicitly training Type~\ref{type:aux} on our data (see Appendix~\ref{app:labor_t}). 
Most remarkably, \textbf{Type~\ref{type:module} (STOP) dominates all other paradigms} in both effectiveness and efficiency. 
For instance, on the AIME 24 benchmark (1.5B), STOP increases average accuracy from 30.10\% to \textbf{37.92\%}—significantly exceeding Type~\ref{type:aux} (32.50\%) and Type~\ref{type:stats} (32.92\%)—while simultaneously reducing total token consumption by over \textbf{73\%}.

\begin{findings}
Type~\ref{type:module} pruning offers better efficiency-accuracy trade-off.
\end{findings}

\subsection{On the Scalability of Pruning}
\label{sec:Scaling}

\begin{figure*}[t]
\vspace{-12mm}
  \centering
  \includegraphics[width=\linewidth]{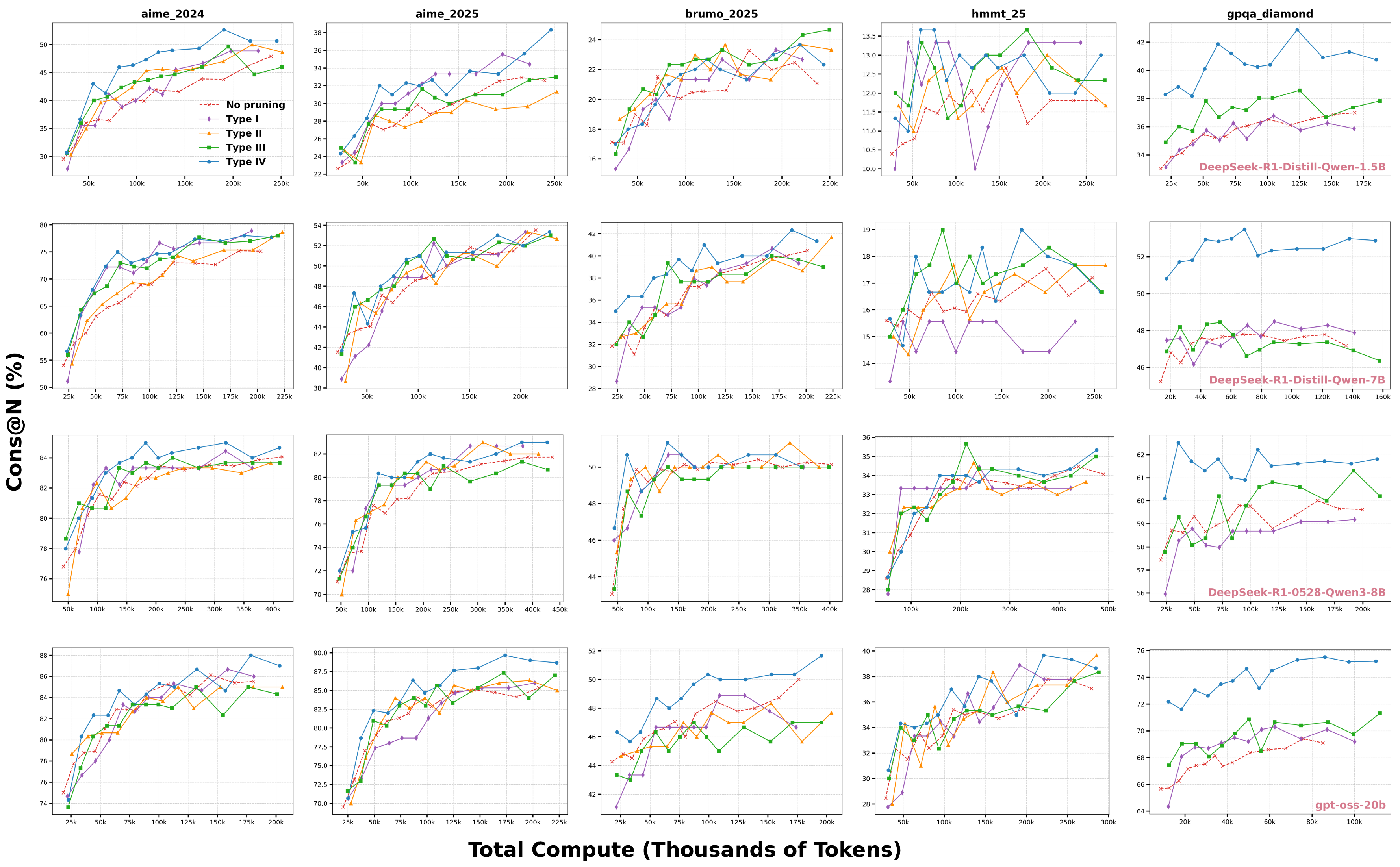}
  \caption{
    Performance vs. compute for four types of $S$ on math and stem benchmarks.
    }
  \label{fig:scaling_main} 
\vspace{-4mm}
\end{figure*}

After validating the effectiveness, we now put these $S$ into practical parallel inference settings to assess their scalability.
We show the cons@N vs. total compute (tokens) in Figure~\ref{fig:scaling_main}.
We fix the retention ratio at $\gamma = M/N = 1/2$ for all methods and vary the initial sample size $N$ to cover different compute budgets.
All other configurations remain consistent with Section~\ref{subsec:On the Effectiveness of Pruning}.

\paragraph{Robustness across Tasks and Model Scales}
We observe a key phenomenon: across all tasks and model scales, some pruning signals achieve better performance than the no-pruning baseline.
However, most existing methods do not exhibit consistent improvements across different tasks and models.
For example, Type~\ref{type:stats} outperforms the baseline on AIME 2024 with the 1.5B model but falls below it on AIME 2025.
In contrast, our proposed Type~\ref{type:module} demonstrates stable and consistently superior scalability across nearly all tasks.
We attribute this robustness to the fact that Type~\ref{type:module} captures the \textbf{intrinsic logical consistency} of reasoning paths, which we further analyze in Section~\ref{subsec:interpretability}.


\begin{findings}\label{findings:XXX}
Type~\ref{type:module} pruning scales robustly across varying compute budgets.
\end{findings}

\section{A Closer Look at STOP}
\subsection{Determining the Optimal remaining ratios}
\label{subsec:Power-law}
While the effectiveness of 
Type~\ref{type:module} is established, optimal deployment requires precise tuning of two critical hyperparameters: the prefix length ($L_\text{prefix}$) and the retention ratio ($\gamma$).
Since increasing $L_\text{prefix}$ generally enhances error detection at the cost of higher latency, users typically fix this parameter according to their specific latency budget.
However, determining the optimal retention ratio $\gamma$ remains non-trivial.
To provide a practical guideline, we formalize the objective as finding a function $\gamma = f(C, L_\text{prefix}, L_\text{task})$ that maximizes accuracy given a compute budget $C$ (in tokens) and a reference task length $L_\text{task}$:
\begin{equation}\small
    \underset{f}{\arg \max} \text{ Accuracy}(C,L_\text{prefix},L_\text{task},\gamma),
\end{equation}
\vspace{-2mm}
where $\gamma$ determines the proportion of paths retained.
Identifying this function $f$ enables the prediction of the optimal $\gamma$ for any given configuration.

\begin{figure*}[ht]
\vspace{-12mm}
  \centering
  \begin{subfigure}[b]{0.24\linewidth}
    \centering
    \includegraphics[width=\linewidth]{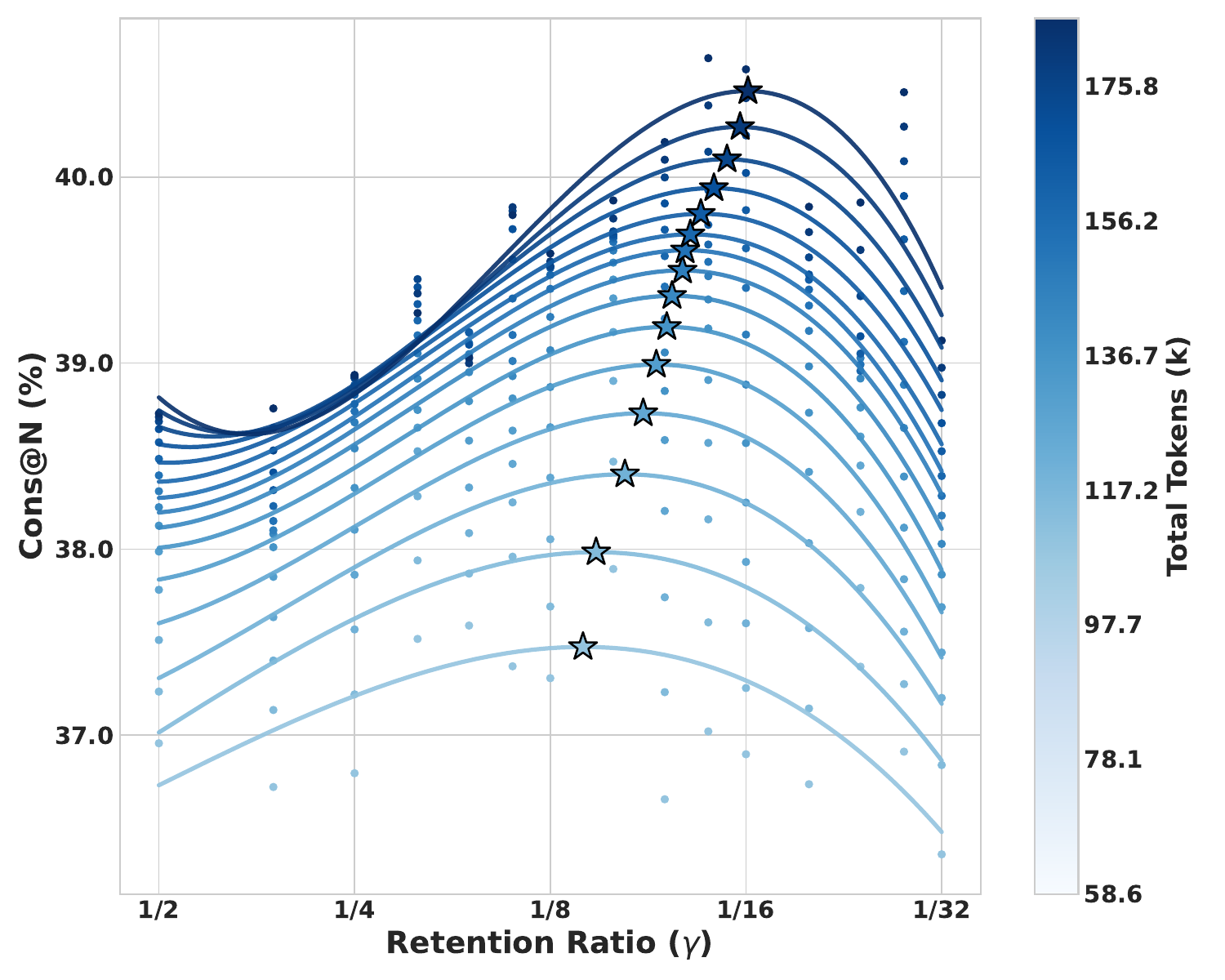}
    \caption{GPQA ($L_{\text{prefix}}=512$)}
  \end{subfigure}
  \hfill
  \begin{subfigure}[b]{0.24\linewidth}
    \centering
    \includegraphics[width=\linewidth]{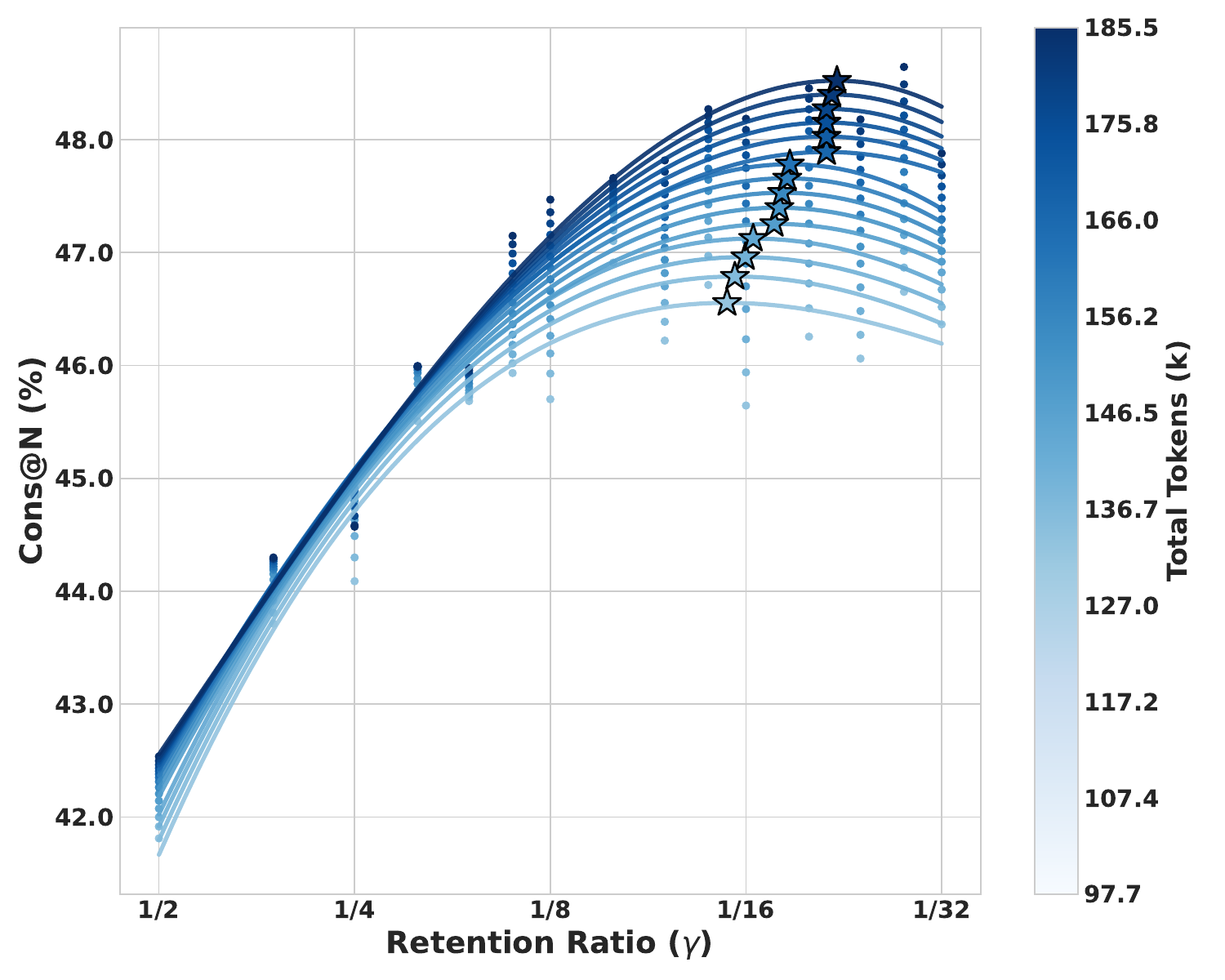}
    \caption{GPQA ($L_{\text{prefix}}=1024$)}
  \end{subfigure}
  \hfill
  \begin{subfigure}[b]{0.24\linewidth}
    \centering
    \includegraphics[width=\linewidth]{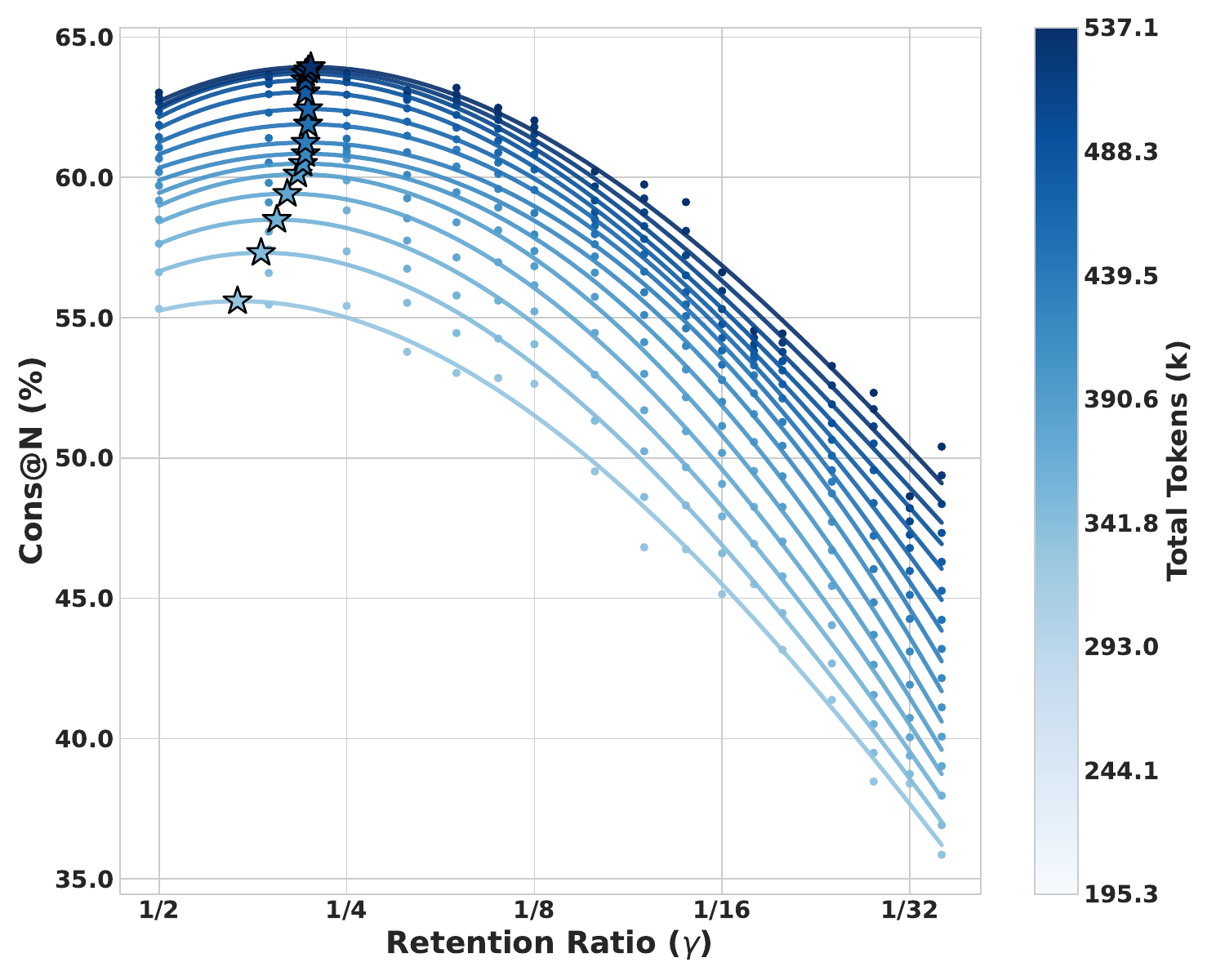}
    \caption{AIME ($L_{\text{prefix}}=2048$)}
  \end{subfigure}
  \hfill
  \begin{subfigure}[b]{0.24\linewidth}
    \centering
    \includegraphics[width=\linewidth]{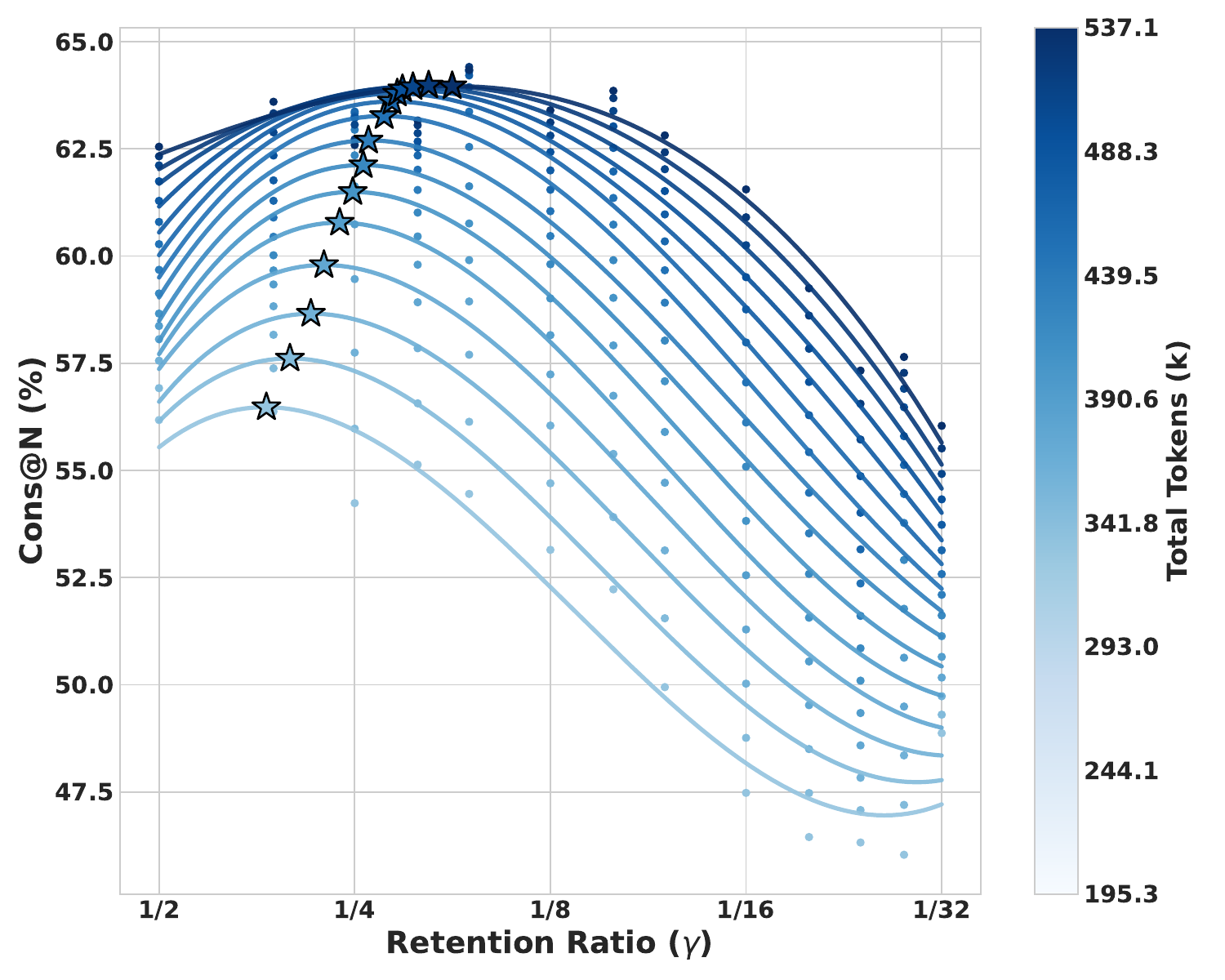}
    \caption{AIME ($L_{\text{prefix}}=4096$)}
  \end{subfigure}
  \caption{
    Performance comparison under different retention ratios ($\gamma$) and prefix lengths ($L_{\text{prefix}}$).
  }
  \label{fig:gamma_all_four}
\vspace{-4mm}
\end{figure*}

\paragraph{Consistent Empirical Trends across Various Settings}
To derive $f$, we conduct experiments using DS-Qwen-2.5-1.5B on AIME 2024 and GPQA Diamond, sweeping $\gamma$ from $1/32$ to $1/2$ across four distinct $L_\text{prefix}$ settings.
The results, plotted in Figure~\ref{fig:gamma_all_four}, exhibit consistent trends: the optimal $\gamma$ decreases as either the compute budget $C$ or the prefix length $L_\text{prefix}$ increases.
These observations indicate that with sufficient compute or richer context, the model identifies futile paths more reliably, thereby allowing for more aggressive pruning (lower $\gamma$) without compromising accuracy.

\begin{figure}[t]
  \centering
  \includegraphics[width=\columnwidth]{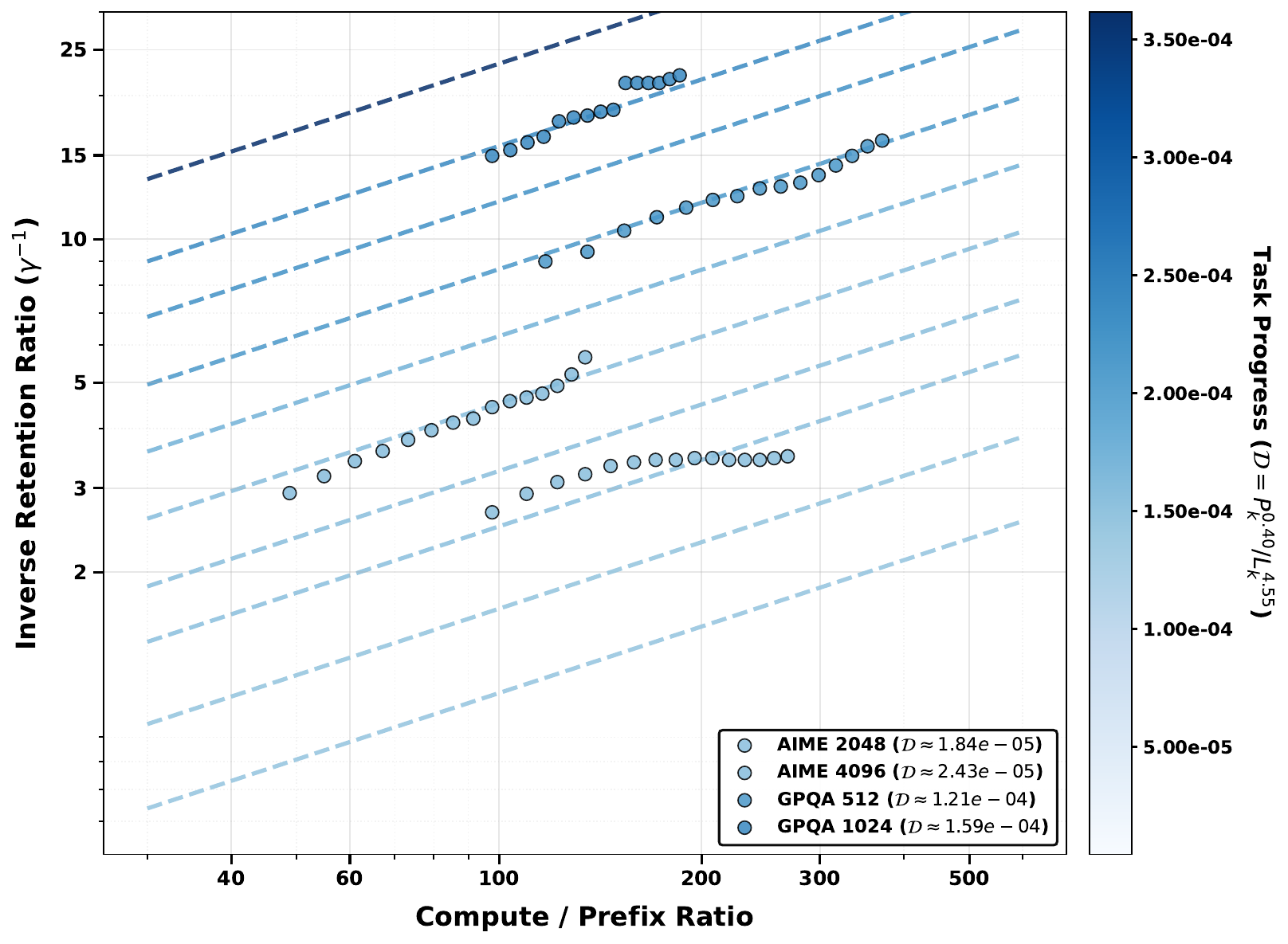}
\caption{
 Inverse retention ratio $\gamma^{-1}$ vs.\ compute-to-prefix ratio. The theoretical curves (Eq.~\ref{eq:optimial_gamma}) closely align with empirical observations across varying reasoning progress levels.
}\label{fig:optimal_pruning_ratio_with_fit}
\vspace{-4mm}
\end{figure}

\paragraph{Formalizing Empirical Findings}
Building on these insights, we model the relationship using a power-law formulation:
\begin{equation}\small
    \gamma^{-1} = f(C, L_{\text{prefix}}, L_{\text{task}}) = aC^b\frac{L_\text{prefix}^c}{L_\text{task}^d}.
    \label{eq:optimial_gamma}
\end{equation}
In this formulation, all input variables are normalized to units of 1,024 tokens.
Fitting this model to our empirical data yields empirical coefficients $a \approx 1.17 \times 10^4$, $b \approx 0.46$, $c \approx 0.40$, and $d \approx 4.55$.
As illustrated in Figure~\ref{fig:optimal_pruning_ratio_with_fit}, the predicted curve aligns closely with the empirical optimal points, offering a robust guideline for parameter selection in practical deployments.

\paragraph{Applying the Empirical Guideline}
To facilitate practical deployment, we apply the derived guideline to predict the optimal retention ratio $\gamma$ for specific configurations without exhaustive search.
Specifically, for a task with a shorter response horizon ($L_{\text{task}} \approx 8{,}650$), a prefix length of $L_{\text{prefix}} = 2{,}048$, and a total compute budget of $C = 158\text{k}$ tokens, the scaling law predicts an optimal inverse retention ratio of $\gamma^{-1} \approx 9.63$, corresponding to $\gamma \approx 10\%$.
Conversely, for a task with a longer reasoning chain ($L_{\text{task}} \approx 12\text{k}$, $L_{\text{prefix}} = 3\text{k}$, and $C = 275\text{k}$), it yields a more conservative estimate of $\gamma^{-1} \approx 3.36$.

These predictions are consistent with our empirical observations, indicating that the scaling law naturally adapts to variations in task complexity.
For detailed lookup guidelines across a broader range of configurations, we refer readers to \textbf{Appendix~\ref{app:gamma_guidelines}}.

\subsection{Ablations and Analysis}\label{subsec:ablation}
To validate the core design choices of \textbf{STOP}, we examine two critical dimensions: the quality of the supervision signal and the computational overhead during inference.

\paragraph{Ablation: Quality of the Supervision Signal}
\textbf{STOP} uses Monte Carlo (MC) estimation with $K=32$ samples to generate probabilistic soft labels ($s^{mc}$), and we compare this setting with binary hard-label supervision, which corresponds to a single-sample estimate ($K=1$). While hard labels are computationally cheap, they introduce high variance because prefix quality depends on a single stochastic continuation. As shown in Table~\ref{tab:ablation_supervision}, increasing the sampling budget from $K=1$ to $K=32$ consistently improves performance. On AIME 2024, soft supervision improves Cons@N from 46.67\% to 53.33\%. These results indicate that MC-based soft labels provide a low-variance signal that enables the lightweight \textbf{STOP} module to learn stable pruning boundaries.

\begin{table}[ht]
\centering

\caption{Performance comparison between hard labels ($K=1$) and MC-estimated soft labels ($K=32$).}
\label{tab:ablation_supervision}
\resizebox{0.9\linewidth}{!}{
\begin{tabular}{l l c c}
\toprule
\textbf{Dataset} & \textbf{Supervision Type} & \textbf{avg@8|64 (\%)} & \textbf{Cons@N (\%)} \\
\midrule
\multirow{2}{*}{AIME 24} & Hard Labels ($K=1$) & 35.42 & 46.67 \\
 & \textbf{Soft Labels ($K=32$)} & \textbf{36.67} & \textbf{53.33} \\
\midrule
\multirow{2}{*}{GPQA} & Hard Labels ($K=1$) & 40.78 & 47.98 \\
 & \textbf{Soft Labels ($K=32$)} & \textbf{41.73} & \textbf{48.48} \\
\bottomrule
\end{tabular}
}
\vspace{-4mm}
\end{table}

\begin{findings}
When training pruning method, soft labels (0.0 to 1.0) have lower variance than hard labels (0 or 1).
\end{findings}

\paragraph{Ablation: Necessity of Critique Adapter}
Given that the LRM's internal states already encode rich reasoning history, a natural question arises: Is a simple linear classifier sufficient to decode the pruning signal?
As shown in Table~\ref{tab:ablation_lora}, the answer is negative. Removing the LoRA adapter leads to a significant performance drop (e.g., from \textbf{36.67\%} to \textbf{31.67\%} on AIME 2024).
This phenomenon highlights a fundamental misalignment: the LRM's native representations are optimized for predicting next token, not value discrimination.
A linear head alone struggles to extract quality assessments from this generation-centric feature space.

\begin{table}[ht]
\centering
\caption{Comparing the STOP module with a simple linear classifier confirms that raw internal states require adaptation to perform effective self-evaluation.}
\label{tab:ablation_lora}
\resizebox{0.9\linewidth}{!}{
\begin{tabular}{l l c c}
\toprule
\textbf{Dataset} & \textbf{Configuration} & \textbf{avg@8|64 (\%)} & \textbf{Cons@N (\%)} \\
\midrule
\multirow{2}{*}{AIME 24} & STOP w/o Adapter & 31.67 & 46.67 \\ 
 & \textbf{STOP} & \textbf{36.67} & \textbf{53.33} \\ 
\midrule
\multirow{2}{*}{GPQA} & STOP w/o Adapter & 33.96 & 35.35 \\ 
 & \textbf{STOP} & \textbf{41.73} & \textbf{48.48} \\ 
\bottomrule
\end{tabular}
}
\vspace{-4mm}
\end{table}

\begin{findings}
High-quality self-correction cannot be achieved by merely probing the states in LRMs; it requires a specialized transformation to bridge the gap between thinking forward (generation) and looking back (reflection).
\end{findings}

\paragraph{Ablation: Sensitivity to Design Choices}
We further examine the sensitivity of \textbf{STOP} to key design choices, namely the number of \texttt{[STOP]} tokens and the LoRA rank. As shown in Table~\ref{tab:ablation_stop_tokens}, performance improves with more tokens, peaks at 4--6, and then degrades with further increases, indicating a trade-off between expressive capacity and overfitting. Similarly, Table~\ref{tab:ablation_rank} shows that moderate ranks (e.g., $r=128$) achieve the best performance, while larger ranks lead to slight degradation, suggesting that excessive capacity is unnecessary.

\begin{findings}
STOP is robust to reasonable hyperparameter choices and does not require large adapters to perform effectively.
\end{findings}

\begin{table}[ht]
\centering
\caption{Effect of the number of \texttt{[STOP]} tokens (DS-Qwen-2.5-1.5B, AIME 2024, $L_{\text{prefix}}=2048$).}
\label{tab:ablation_stop_tokens}
\resizebox{0.6\linewidth}{!}{
\setlength{\tabcolsep}{4pt}
\begin{tabular}{c c | c c}
\toprule
\# Tokens & avg@32$|$256 & \# Tokens & avg@32$|$256 \\
\midrule
1 & 30.10 & 6 & \textbf{37.71} \\
2 & 33.54 & 7 & 36.15 \\
3 & 35.94 & 8 & 35.00 \\
4 & 36.86 & 9 & 33.65 \\
5 & 36.77 & - & - \\
\bottomrule
\end{tabular}
}
\vspace{-5mm}
\end{table}

\begin{table}[ht]
\vspace{-2mm}
\centering
\caption{Effect of LoRA rank (DS-Qwen-2.5-1.5B, AIME 2024).}
\label{tab:ablation_rank}
\resizebox{0.5\linewidth}{!}{
\setlength{\tabcolsep}{4pt}
\begin{tabular}{c c c}
\toprule
Rank & Params (M) & avg@8$|$64 \\
\midrule
32  & 36.9  & 32.50 \\
64  & 73.9  & 36.25 \\
\textbf{128} & \textbf{147.7} & \textbf{36.67} \\
256 & 295.4 & 35.83 \\
\bottomrule
\end{tabular}
}
\vspace{-5mm}
\end{table}

\paragraph{Analysis: Computational Overhead}
We quantify the inference latency on a single NVIDIA H100 GPU using DS-Qwen-2.5-7B with a fixed prefix length of $2,048$.
As detailed in Table~\ref{tab:latency}, existing paradigms incur notable costs: Type~\ref{type:aux} requires full sequence re-encoding, resulting in the highest latency (\textbf{1.13\,s}, 3.37\% overhead), while Type~\ref{type:heuristic} suffers from the computational bottleneck of pairwise similarity calculations (\textbf{0.38\,s}).
In stark contrast, \textbf{STOP} (Type~\ref{type:module}) minimizes overhead to a negligible \textbf{0.20\,s} (\textbf{0.59\%}).
This efficiency stems directly from our architectural design: by \textbf{reusing the pre-computed KV cache} and restricting verification to a single forward pass of special tokens, STOP eliminates redundant computation, ensuring high-throughput deployment.

\begin{table}[ht]
\centering
\caption{Inference overhead analysis. STOP achieves near-zero cost by avoiding re-encoding.}
\label{tab:latency}
\resizebox{0.85\linewidth}{!}{
\begin{tabular}{l c c}
\toprule
\textbf{Pruning Paradigm} & \textbf{Latency / Check} & \textbf{Relative Overhead} \\
\midrule
Type~\ref{type:aux} & 1.13\,s & 3.37\% \\
Type~\ref{type:heuristic} & 0.38\,s & 0.93\% \\
\rowcolor{gray!10} \textbf{Type~\ref{type:module} (STOP)} & \textbf{0.20\,s} & \textbf{0.59\%} \\
\bottomrule
\end{tabular}
}
\vspace{-4mm}
\end{table}

\paragraph{Analysis: Generalization to Non-Math/STEM Tasks}
To assess whether \textbf{STOP} captures universal reasoning patterns beyond mathematics and science, we extend our evaluation to \textbf{ZebraLogic}, a benchmark designed to evaluate combinatorial reasoning and constraint satisfaction capabilities through logic grid puzzles.
Specifically, we conduct experiments on the multiple-choice mode (\texttt{mc\_mode}) to test reasoning under constraints.
Using the \textbf{DS-Qwen-2.5-7B} model, we evaluate 500 randomly sampled instances of moderate difficulty (Rows, Cols $\le 4$).
As shown in Table~\ref{tab:ablation_zebra}, \textbf{STOP} improves accuracy from 73.73\% to \textbf{77.23\%}.
This consistent gain confirms that the pruning signals learned by the module are not strictly domain-dependent, but rather transferable to general logical inference tasks.

\begin{table}[t]
\centering
\caption{\textbf{Generalization on ZebraLogic.} \textbf{STOP} \textbf{robustly generalizes} beyond math and science tasks.}
\label{tab:ablation_zebra}
\resizebox{0.9\linewidth}{!}{
\begin{tabular}{l c c c}
\toprule
\multirow{2}{*}{\textbf{Model}} & \textbf{No pruning (Baseline)} & \textbf{\textbf{STOP}} & \multirow{2}{*}{\textbf{Gain}} \\
\cmidrule(lr){2-2} \cmidrule(lr){3-3} 
 & \textbf{avg@64 (\%)} & \textbf{avg@8|64 (\%)} & \\
\midrule
DS-Qwen-2.5-7B & 73.73 & \textbf{77.23} & \textbf{+3.50\%} \\
\bottomrule
\end{tabular}
}
\vspace{-3mm}
\end{table}

\paragraph{Analysis: Generalization to Tool Use}
We further evaluate whether \textbf{STOP} generalizes to realistic tool-use scenarios by submitting our system to the \textbf{AIMO3} competition, where models solve mathematical problems with access to external tools under a fixed evaluation protocol. Built on a \textbf{GPT-OSS-120B + tool} framework, we compare against a baseline that directly performs parallel reasoning without pruning under the same resource constraints; due to the competition setting (single H100 GPU and a 5-hour limit for 50 problems), the baseline cannot scale to larger sampling budgets. As shown in Table~\ref{tab:ablation_tool}, both STOP configurations consistently outperform the baseline, improving the score from \textbf{39} to \textbf{42} (24$\rightarrow$8) and \textbf{43} (16$\rightarrow$8), with the best configuration reaching \textbf{silver-level performance} on the public leaderboard, demonstrating that \textbf{STOP} remains effective in tool-augmented reasoning and translates into tangible gains in real-world competitive settings.



\begin{table}[h]
\centering
\small
\setlength{\tabcolsep}{5pt}
\caption{Results on the AIMO3 competition setting with tool use (GPT-OSS-120B).}
\label{tab:ablation_tool}
\begin{tabular}{l c}
\toprule
\textbf{Method} & \textbf{Score} \\
\midrule
Baseline + Tool & 39 \\
STOP (24$\rightarrow$8) & 42 \\
STOP (16$\rightarrow$8) & \textbf{43} \\
\bottomrule
\end{tabular}
\vspace{-3mm}
\end{table}

\subsection{How STOP Attends}
\label{subsec:interpretability}
\begin{figure*}[t]
\vspace{-14mm}
  \centering
  \begin{subfigure}[b]{0.48\linewidth}
    \centering
    \includegraphics[height=5.0cm, keepaspectratio]{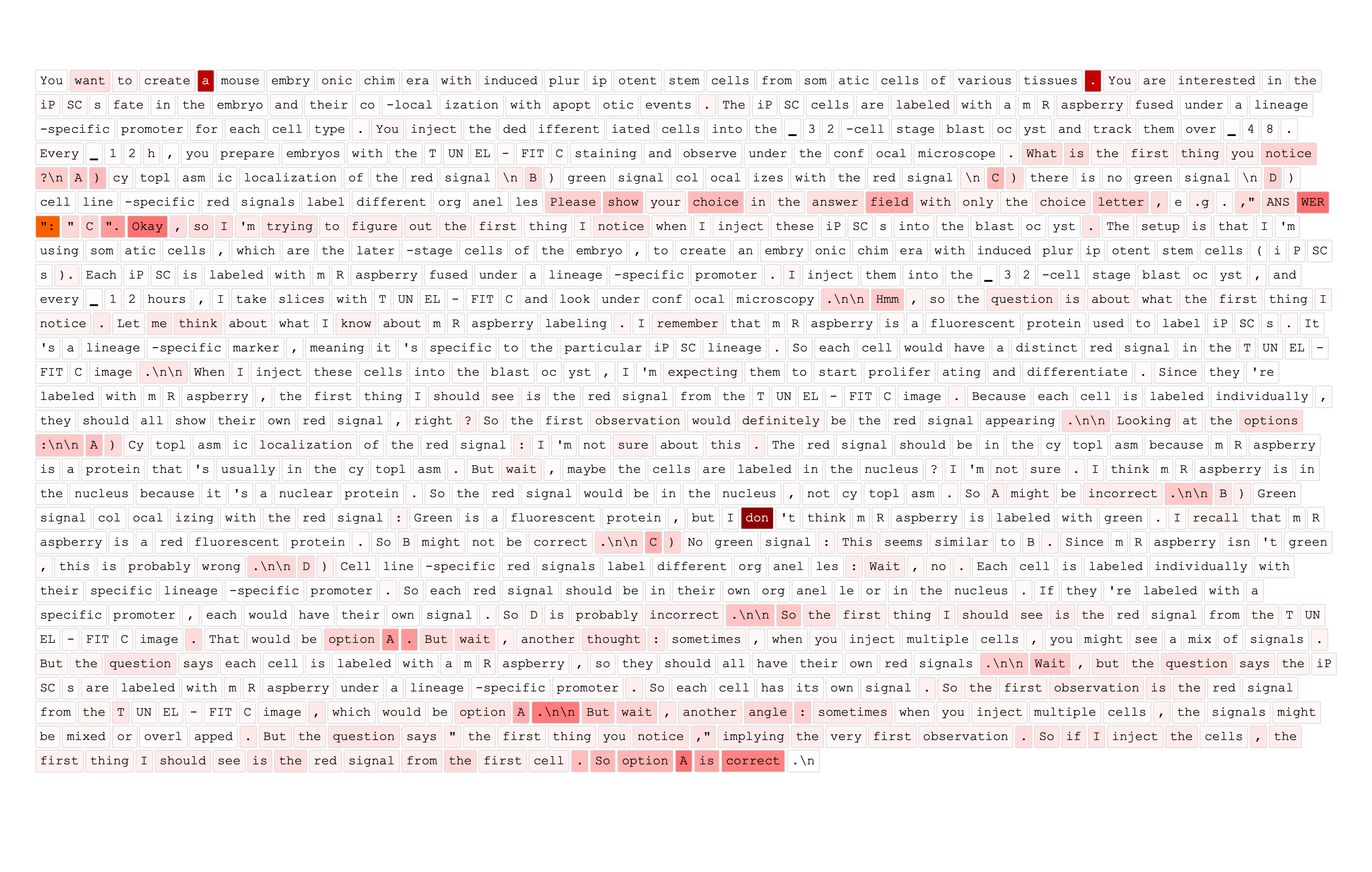}
    \caption{\textbf{High-scoring Path}}
    \label{fig:attn_correct}
  \end{subfigure}
  \hfill
  \begin{subfigure}[b]{0.48\linewidth}
    \centering
    \includegraphics[height=5.0cm, keepaspectratio]{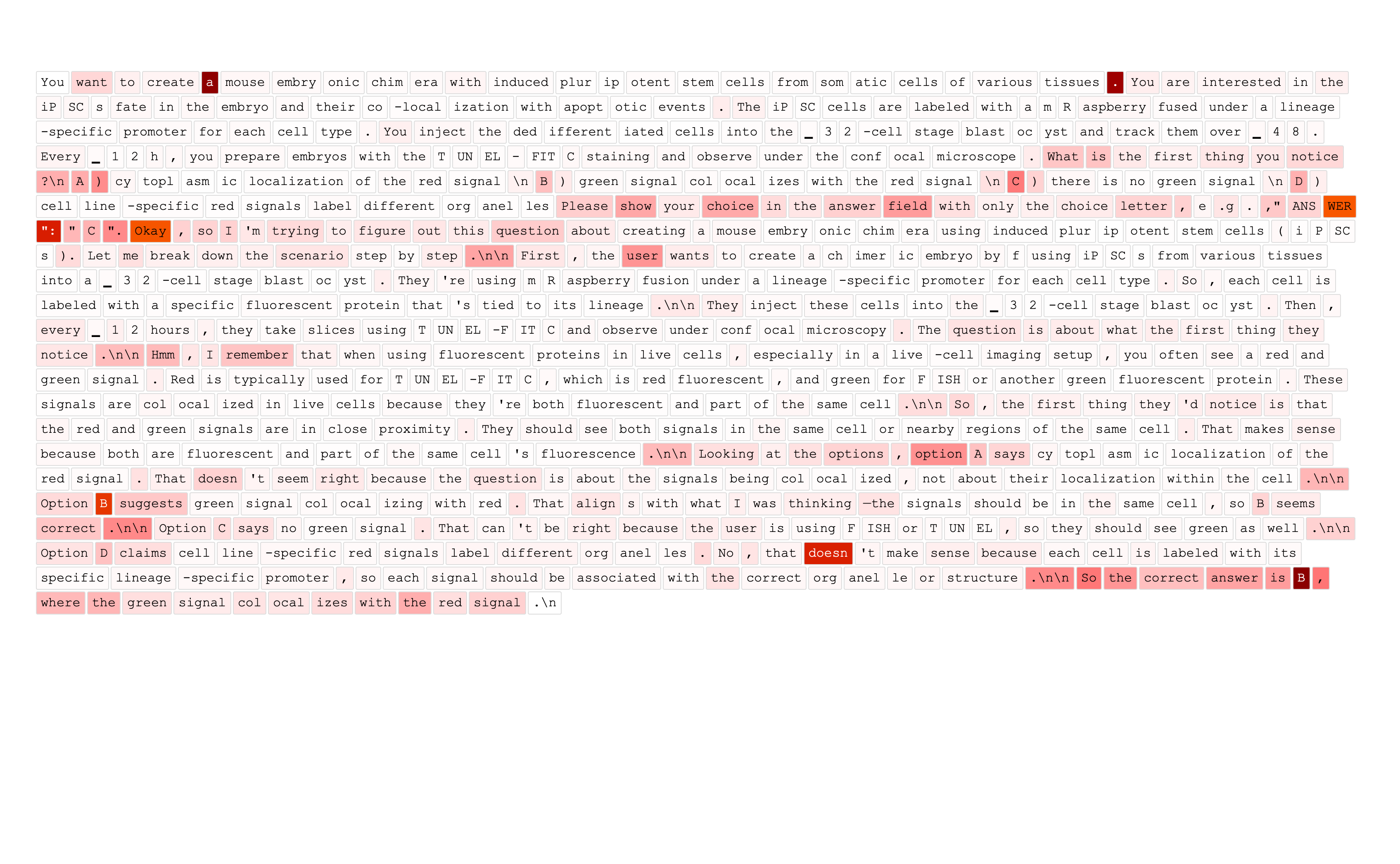} 
    \caption{\textbf{Low-scoring Path}}
    \label{fig:attn_incorrect}
  \end{subfigure}
  \vspace{-4mm}
\caption{\textbf{Attention Analysis of \texttt{[STOP]} Decision-Making.} High-scoring paths prioritize logical pivots (e.g., self-correction markers), whereas low-scoring paths fixate on terminal answer tokens. This contrast confirms that \textbf{STOP} functions as a process-oriented evaluator, rewarding reasoning integrity over premature closure.}
\label{fig:attention_analysis}
  \vspace{-5.0mm}
\end{figure*}

To understand how \textbf{STOP} distinguishes valid reasoning trajectories, we visualize the attention distribution of the \texttt{[STOP]} token (Figure~\ref{fig:attention_analysis}).
Overall, the module exhibits a broad attention pattern. It consistently attends to \textbf{multiple-choice options} (A, B, C, D) as well as discourse markers (e.g., ``Hmm'', ``Wait''), which enables it to track the structural progression of the reasoning process.

\paragraph{Process-oriented Evaluation}
Importantly, high-scoring and low-scoring trajectories present clearly distinct attention signatures.
In the \textbf{high-score case} (Figure~\ref{fig:attention_analysis}a), attention prioritizes the reasoning process rather than the final outcome.
Specifically, the \texttt{[STOP]} token focuses on \textbf{cognitive pivots} (e.g., the negation ``don't''), indicating an emphasis on logical operations that trigger self-correction.
In contrast, the \textbf{low-score case} (Figure~\ref{fig:attention_analysis}b) demonstrates a pattern of \textbf{premature closure}: attention shifts early to the terminal token (e.g., ``B'') while critical logical markers receive little attention.
Consequently, \textbf{STOP} penalizes such trajectories and interprets the lack of attention to logical pivots as evidence of reasoning failure.
See Appendix~\ref{app:mechanistic_interpretability} for more cases.

\section{Conclusion}
In this work, we address the critical efficiency bottleneck of parallel reasoning by establishing the first unified taxonomy of path pruning. 
This framework not only resolves the fragmentation in existing research but also reveals the unexplored potential of \textit{learnable internal methods} (Type~\ref{type:module}). 
To bridge this gap, we introduce \textbf{STOP}, a lightweight method that leverages internal representations to identify and terminate futile prefixes effectively. 
Extensive evaluations demonstrate that STOP consistently dominates existing paradigms, significantly enhancing reasoning accuracy while reducing token consumption by over 70\%.
Moreover, we resolve scalability and deployment uncertainties by deriving a robust interaction formulation. 
This provides practitioners with a precise empirical guideline for optimizing the trade-off between exploration and exploitation under varying computational constraints.
Finally, our in-depth analysis of the mechanism and architectural choices offers valuable insights to guide future research.

\section*{Acknowledgment}
We respectfully note that OTV~\cite{zhuang2026one} is our concurrent work, and they are also the first Type~\ref{type:module} method.

This work was supported by Major Frontier Exploration Program (Grant No. C10120250085) from the Shenzhen Medical Academy of Research and Translation (SMART),  Shenzhen Medical Research Fund (B2503005),  NSFC grant 72495131, the 1+1+1 CUHK-CUHK(SZ)-GDSTC Joint Collaboration Fund, Guangdong Provincial Key Laboratory of Mathematical Foundations for Artificial Intelligence (2023B1212010001), and the International Science and Technology Cooperation Center, Ministry of Science and Technology of China (under grant 2024YFE0203000).

\section*{Limitations}
As the pioneering instantiation of the internal learnable paradigm (Type~\ref{type:module}), \textbf{STOP} validates the potential of intrinsic representations for trajectory pruning. 
However, we acknowledge specific limitations in our current scope and highlight promising directions for future research.

\paragraph{Limitations.}
\begin{itemize}
    \item \textbf{Verification at Extreme Scales} Our current evaluation spans models up to 20B parameters and standard compute budgets (e.g., $N=64$). The behavior of STOP on substantially larger models (e.g., 70B+) and under massive sampling regimes (e.g., $N \ge 1000$) remains to be empirically verified.
    \item \textbf{Structural Flexibility} This work focuses on single-stage pruning at fixed positions (e.g., $L_{\text{prefix}}=2048$). We have not yet explored more complex settings, such as multi-stage sequential pruning or unstructured pruning where checkpoints are determined dynamically rather than at fixed token indices.
\end{itemize}

\paragraph{Future Directions.}
\begin{itemize}
    \item \textbf{Progressive Multi-Stage Pruning} A natural extension is to apply STOP in a cascading manner (e.g., funneling candidates from $64 \to 32 \to 16$ at successive checkpoints). This "progressive filtering" strategy could further optimize the compute allocation by dynamically narrowing the search space as reasoning deepens.
    \item \textbf{Accelerating RL Training} Beyond inference, STOP holds significant potential for training efficiency. In Reinforcement Learning (e.g., PPO or GRPO), STOP can serve as an online rejection mechanism during the rollout phase, terminating low-value trajectories early to increase the density of high-quality training signals per unit of compute.
\end{itemize}

\bibliography{custom}

\newpage
\appendix

\appendix

\section{Related Work}
\subsection{Parallel Reasoning}
\label{subsec:related_parallel}
Parallel reasoning, which generates multiple trajectories to verify or aggregate answers, has become a standard paradigm for enhancing LRM performance. 
A recent survey by ~\cite{wang2025parallel} systematically categorizes these approaches into three dimensions:
\textbf{(1) Non-interactive Reasoning}, which generates independent paths without communication, including majority voting in \textit{Self-Consistency}~\cite{wang2022self}, ranking in \textit{Best-of-N}~\cite{brown2024large}, and structured exploration in \textit{Tree-of-Thoughts}~\cite{yao2023tree}.
\textbf{(2) Interactive Reasoning}, which enables active information exchange among paths, for example, internal state sharing in \textit{Leap}~\cite{luo2025peers} or multi-agent collaboration~\cite{chan2023chamele}.
\textbf{(3) Efficiency Optimization}, which focuses on accelerating decoding mechanics, such as speculative decoding in \textit{Medusa}~\cite{cai2024medusa}.
Although these methods enhance reasoning performance, they still suffer from substantial inference costs, which remain a major limitation.


\subsection{Path Pruning (Prefix Rejection)}
To mitigate the high inference cost of parallel reasoning, path pruning strategies aim to terminate unpromising trajectories early. Consistent with the taxonomy in Section~\ref{sec:taxonomy}, we categorize existing works based on signal source and learnability.

Regarding \textbf{external} signals, \textbf{non-learnable} methods (Type~\ref{type:heuristic}) like SlimSC~\cite{hong2025slimsc} prune paths utilizing heuristic metrics such as semantic similarity to minimize redundancy.
In contrast, \textbf{learnable} approaches (Type~\ref{type:aux}) rely on trained verifiers. This category encompasses discriminative classifiers used in DeepPrune~\cite{tu2025deepprune} and LaBoR~\cite{liao2025lost}, as well as generative verifiers in ThinkPRM~\cite{khalifa2025process} and multi-agent frameworks like MAV~\cite{lifshitz2025multi}.
Shifting to \textbf{internal} sources, \textbf{non-learnable} methods (Type~\ref{type:stats}) derive signals directly from intrinsic statistics. Representative works include confidence-based estimation in DeepConf~\cite{fu2025deepthink} and AdaDec~\cite{he2025adadec}, or entropy-based metrics in Think Just Enough~\cite{sharma2025thinkjust}.

Notably, \textbf{prior works} leave the quadrant of \textbf{internal learnable} modules (Type~\ref{type:module}) unexplored. \textbf{\textbf{STOP}} is designed to bridge this gap, utilizing a trainable adapter to extract rich internal semantics, thus offering a solution that is both structurally efficient and data-driven.

\begin{figure}[t]
  \centering
  \includegraphics[width=0.8\columnwidth]{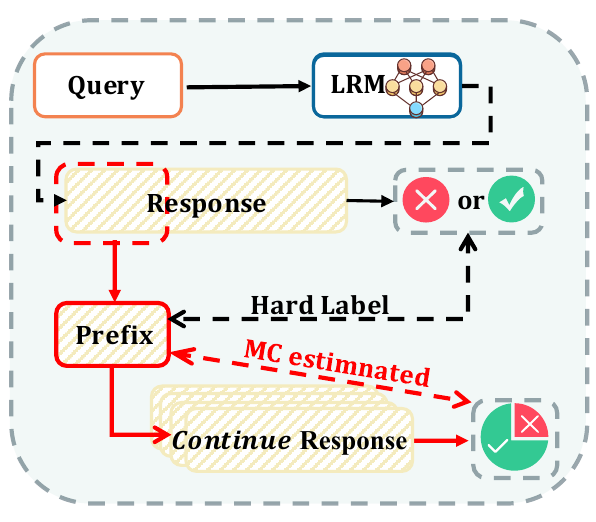}
  \caption{MC-based construction of prefix--potential supervision.}
  \label{fig:Data_Construction}
  \vspace{-2mm}
\end{figure}

\section{Data Construction Details}
\label{app:data_construction}

To train our \textbf{STOP} module, we require a dataset that directly maps prefixes of reasoning paths to the probability that the final answer succeeds. A single binary label on a complete path provides an insufficient and noisy signal, because a promising prefix may still end in an accidental failure, while a flawed prefix may occasionally be recovered by chance. Therefore, we construct a dataset of \texttt{(prefix, success probability)} pairs using \textbf{Monte Carlo (MC) estimation}~\cite{wang2024mathshepherd,zhao2025genprm}.

\subsection{Source Benchmarks and Decontamination}
We constructed a supervised fine-tuning dataset derived from high-quality mathematical and scientific benchmarks. Specifically, we aggregated approximately 1,000 problems from the \textbf{AIME} competition (spanning years 1984 to 2023)~\cite{maa2024aime,maa2025aime}, augmented with the non-Diamond portion of the \textbf{GPQA} dataset~\cite{rein2024gpqa}. 
\textit{Crucially, to ensure zero data leakage, we strictly excluded the evaluation sets from this training corpus: specifically, \textbf{AIME 2024}, \textbf{AIME 2025}, and the \textbf{GPQA Diamond} subset were entirely removed.}

\subsection{Model-Specific Construction Pipeline}
Since reasoning capabilities vary across model scales, we adopted a \textbf{model-specific pipeline} where each LRM (e.g., 1.5B) generates its own training data. The procedure proceeds as follows:

\paragraph{Difficulty Stratification (Filtering).}
Before generating prefixes, we first filter source problems to focus on the model's \textit{learnable boundary}. For each problem, we generate $N=32$ reasoning paths and calculate the pass rate. We explicitly exclude \textbf{trivial samples} ($>28$ correct answers) that the model has already mastered, as well as \textbf{intractable samples} ($<4$ correct answers) likely beyond its current capacity. This ensures that the training data consists of problems where the pruning signal is most valuable.

\begin{table}[t]
\centering
\small
\setlength{\tabcolsep}{4pt}
\caption{\textbf{Statistics of model-specific training data.} Prefixes are extracted from Math (AIME) and Science (GPQA). Data volume decreases for larger models due to filtering of trivial samples.}
\label{tab:training_data_stats}
\begin{tabular}{l c c c}
\toprule
\textbf{Model} & \textbf{Math} & \textbf{Science} & \textbf{Total} \\
\midrule
DS-Qwen-2.5-1.5B & 14,816 & 8,448 & \textbf{23,264} \\
DS-Qwen-2.5-7B   & 12,092 & 5,666 & \textbf{17,758} \\
DS-Qwen-3-8B     & 10,848 & 4,456 & \textbf{15,304} \\
GPT-OSS-20B      & 7,872  & 2,378 & \textbf{10,250} \\
\bottomrule
\end{tabular}
\vspace{-3mm}
\end{table}

\paragraph{Prefix Generation.} 
From the retained problems, we use the LRM to generate a prefix $p$ that forms part of a complete reasoning trajectory. To simulate a realistic mid-generation checkpoint, we truncate these paths at a fixed length of \textbf{$L_{\text{prefix}} = 2{,}048$} tokens.

\paragraph{Potential Estimation via MC Rollouts.}
To estimate the potential of $p$, we fix the prefix and generate \textbf{$K=32$} continuations under a temperature of $0.6$. This procedure produces a set of full-length responses $\{\tau'_{1}, \tau'_{2}, \ldots, \tau'_{K}\}$.

\paragraph{MC Score Calculation.}
We evaluate each response for correctness (1 if correct and 0 otherwise). The MC-estimated success probability $s^{mc}$ is defined as the empirical accuracy:
\begin{equation} \small
    s^{mc} = \frac{1}{K} \sum_{j=1}^{K} \text{is\_correct}(\tau'_{j}).
\end{equation}
The resulting label $s^{mc} \in [0.0, 1.0]$ provides a fine-grained probabilistic target used to train the STOP module.

\paragraph{Data Statistics and Insights.}
Table~\ref{tab:training_data_stats} summarizes the composition of the constructed datasets. 
We observe a distinct \textbf{inverse scaling trend}: as the model size increases, the number of valid training samples decreases (e.g., from 23,264 for 1.5B to 10,250 for 20B). 
This confirms the efficacy of our difficulty stratification strategy: larger models (e.g., GPT-OSS-20B) achieve high pass rates ($>28/32$) on a larger portion of the source benchmarks, causing these ``trivial'' instances to be filtered out. Consequently, the training data naturally adapts to focus on the \textit{learnable boundary} specific to each model's capability.


\subsection{Training Cost Details}
\label{app:training_cost}

Constructing the MC supervision dataset requires sampling multiple continuations per prefix (e.g., $K=32$) as described in Section~\ref{subsec:motivation}.
In practice, we find that moderate sampling budgets provide a good balance between estimation stability and computational cost, as also reflected in our ablation results.
We report the estimated cost across different model scales in Table~\ref{tab:training_cost}.

These costs correspond to a one-time data construction process.
Once constructed, the dataset can be reused across training runs and model variants, amortizing the cost of data construction.
The trained STOP module introduces negligible overhead during inference.
These costs are reported to provide transparency and should be interpreted as approximate estimates depending on implementation and hardware configurations.
\begin{table*}[t]
\centering
\small
\caption{\textbf{Training Cost for MC Supervision Construction.} 
We report the number of training pairs and the estimated wall-clock cost (in 8$\times$H100 GPU hours) required to construct the dataset with $K=32$ Monte Carlo samples per prefix.}
\label{tab:training_cost}
\begin{tabular}{lcccc}
\toprule
\textbf{Model} & \textbf{Math} & \textbf{Science} & \textbf{Total Training Pairs} & \textbf{8$\times$H100 Hours} \\
\midrule
DS-Qwen-2.5-1.5B & 14,816 & 8,448 & \textbf{23,264} & 43.08 \\
DS-Qwen-2.5-7B   & 12,092 & 5,666 & \textbf{17,758} & 39.46 \\
DS-Qwen-3-8B     & 10,844 & 4,456 & \textbf{15,304} & 37.79 \\
GPT-OSS-20B      & 7,872  & 2,378 & \textbf{10,250} & 75.93 \\
\bottomrule
\end{tabular}
\vspace{-3mm}
\end{table*}

\begin{table*}[t]
\centering
\small
\caption{\textbf{Training hyperparameters across model scales.}}
\label{tab:training_hyperparams}
\begin{tabular}{lcccc}
\toprule
\textbf{Hyperparameter} & \textbf{1.5B} & \textbf{7B} & \textbf{8B} & \textbf{20B} \\
\midrule
Per-Device Batch Size & 16 & 8 & 8 & 2 \\
Gradient Accumulation & 1 & 2 & 2 & 8 \\
Learning Rate & $2 \times 10^{-5}$ & $2 \times 10^{-5}$ & $2 \times 10^{-5}$ & $2 \times 10^{-5}$ \\
LoRA Rank ($r$) & 128 & 256 & 256 & 2048 \\
LoRA Alpha ($\alpha$) & 256 & 512 & 512 & 4096 \\
Target Modules & All Linear & All Linear & All Linear & All Linear \\
Optimizer & AdamW & AdamW & AdamW & AdamW \\
Max Prefix Length & 2048 & 2048 & 2048 & 2048 \\
Training Epochs & 15 & 15 & 15 & 15 \\
Precision & \texttt{bf16} & \texttt{bf16} & \texttt{bf16} & \texttt{bf16} \\
\bottomrule
\end{tabular}
\vspace{-3mm}
\end{table*}

\section{Detailed Experimental Settings}
\label{app:exp_setup}

In this appendix, we provide the complete experimental details to ensure reproducibility, covering infrastructure, datasets, input formats, training hyperparameters, and baseline implementations.

\subsection{Infrastructure and Sampling Configuration}
\label{app:infrastructure}
\noindent\textbf{Infrastructure.}
All experiments were conducted on NVIDIA H100 (80GB) GPUs.
We utilized the \texttt{vLLM} framework~\cite{kwon2023efficient} to support efficient batched inference during the evaluation phases.

\noindent \textbf{Sampling Configuration.}
To ensure consistency across all pruning methods, we adopted a unified generation configuration.
Specifically, the temperature was set to $0.6$, top-$p$ to $0.95$, and top-$k$ to $40$.
The maximum generation length was set to $16{,}384$ tokens for the 1.5B and 7B models, and $32{,}768$ tokens for the 8B and 20B models.
For \texttt{gpt-oss} models, the reasoning effort was set to ``medium''.

\subsection{Evaluation Protocol}
\label{app:datasets}


We strictly adhered to established evaluation protocols to ensure fair comparison and reproducibility. The \textbf{GPQA-Diamond} subset, consisting of 198 high-difficulty questions, was reserved exclusively as a held-out test set. Consequently, all remaining GPQA questions were used solely during the training stage. This rigorous separation guarantees zero information leakage from the training corpus to the evaluation benchmarks.

\subsection{Prompt Templates and Input Format}
\label{app:prompts}

To ensure rigorous reproducibility, we detail the exact prompt templates and input construction used in our experiments. We utilized the standard zero-shot Chain-of-Thought (CoT) format.

\begin{tcolorbox}[
    colback=white, 
    colframe=gray!60!black, 
    title=\textbf{Prompt Templates \& Input Format},
    fonttitle=\bfseries\large,
    boxrule=1pt,
    arc=2mm
]
\noindent\textbf{\textcolor{blue}{GPQA:}}

\noindent\texttt{Please show your choice in the answer field with only the choice letter, e.g., "ANSWER": "C".}

\vspace{1em} 

\noindent\textbf{\textcolor{blue}{Math Tasks (AIME, HMMT, BRUMO):}}

\noindent\texttt{Please reason step by step, and put your final answer within \textbackslash boxed\{\}.}

\vspace{1em}
\tcbline 
\vspace{0.5em}



\noindent\textbf{STOP Module Input Mechanism:}

To achieve zero-overhead verification, we do not re-encode the full text. Instead, the \textbf{STOP} token is appended directly to the \textbf{pre-computed KV cache} of the generated prefix.
Conceptually, this provides the module with the following effective context, allowing it to attend to the full history:

\begin{center}
\fcolorbox{gray!60}{gray!10}{
\parbox{0.9\columnwidth}{
\centering\ttfamily
[User Prompt] [Generated Reasoning Prefix] [STOP]
}}
\end{center}

\end{tcolorbox}

\subsection{STOP Module Training Details}
\label{app:training_details}

We developed a custom training pipeline utilizing the Hugging Face \texttt{Accelerate} and \texttt{PEFT} libraries. All experiments were conducted on 8 NVIDIA H100 GPUs using a LoRA-only approach. We froze the base model parameters and strictly trained low-rank adapters attached to \textbf{all linear layers} within the transformer blocks. 
Specifically, we targeted the full set of projections: \texttt{q\_proj}, \texttt{k\_proj}, \texttt{v\_proj}, \texttt{o\_proj}, \texttt{gate\_proj}, \texttt{up\_proj}, and \texttt{down\_proj}. 
The specific hyperparameters, including the varying LoRA configurations for different model scales, are detailed in Table~\ref{tab:training_hyperparams}.


\subsection{Baseline Descriptions}
\label{app:baselines}
We provide additional details on the baseline implementations used in Section~\ref{sec:Comprehensive}:
\begin{itemize}
    \item \textbf{SlimSC~\cite{hong2025slimsc} (Type~\ref{type:heuristic}):} Computes the pairwise Jaccard similarity between the current generation and previously explored reasoning paths. It prunes trajectories that exhibit high semantic redundancy to ensure diversity.
    \item \textbf{LaBoR~\cite{liao2025lost} (Type~\ref{type:aux}):} Relies on a separate, trained Process Reward Model (PRM) to score generated prefixes. We used the official checkpoints released by the authors where available.
    \item \textbf{DeepConf~\cite{fu2025deepthink} (Type~\ref{type:stats}):} Estimates confidence by computing perplexity and entropy directly from the model logits of the generated tokens, serving as a non-learnable internal baseline.
\end{itemize}

\section{Ablation: Data Quality vs. Architecture}
\label{app:labor_t}

\subsection{Motivation and Setup}
A potential confounding factor in our main results is the quality of the training data. Since \textbf{STOP} is trained on a high-quality dataset constructed via Monte Carlo rollouts, it is natural to hypothesize that the observed performance gains mainly arise from superior supervision rather than from the \textbf{Type~\ref{type:module}} architecture itself. To disentangle these two factors, we introduce a controlled baseline, \textbf{Type~\ref{type:aux}${}^{\text{retrain}}$ (Retrained Early Pruning)}. LaBoR~\cite{liao2025lost} propose an Early Pruning strategy based on an external Process Reward Model (PRM), specifically \texttt{Qwen2.5-Math-PRM-7B}, but their model is not trained on our MC-estimated soft labels. For a fair comparison, we adopt the same architecture and fine-tune it on the \emph{same dataset} of prefix--success probability pairs used to train \textbf{STOP}. This comparison isolates the architectural effect between an internal, learnable method (\textbf{Type~\ref{type:module}}) with access to full hidden states and an external reward model (\textbf{Type~\ref{type:aux}}) that relies only on token-level outputs, thereby ruling out data quality as the sole source of improvement. \textbf{Note:} Because the backbone of Type~\ref{type:aux} is specialized for mathematics, we exclude the GPQA (Science) benchmark from this ablation, as the external PRM lacks sufficient domain knowledge for scientific reasoning.

\begin{table*}[t]
\centering
\caption{
\textbf{Ablation Study: Architecture vs. Data.} Comparison of avg@8 and token efficiency. 
\textbf{Type~\ref{type:aux}} refers to the standard external PRM baseline (Early Pruning). 
\textbf{Type~\ref{type:aux}${}^{\text{retrain}}$} denotes the same external architecture retrained on our MC-estimated data. 
\textbf{STOP} (\textbf{Type~\ref{type:module}}) outperforms both, demonstrating that architectural access to internal states yields gains beyond data quality alone.
Note: Type~\ref{type:aux} variants are not evaluated on GPQA due to the domain limitation of the math-specialized PRM backbone.
}
\label{tab:main_results_LaBor}
\resizebox{\textwidth}{!}{%
\begin{tabular}{llcc cc cc cc}
\toprule
\multirow{2}{*}{\textbf{Model}} & \multirow{2}{*}{\textbf{Dataset}} 
& \multicolumn{2}{c}{\textbf{Full Paths (Baseline)}} 
& \multicolumn{2}{c}{\textbf{Type~\ref{type:aux}}} 
& \multicolumn{2}{c}{\textbf{Type~\ref{type:aux}${}^{\text{retrain}}$}} 
& \multicolumn{2}{c}{\textbf{Type~\ref{type:module}}} \\
\cmidrule(lr){3-4} \cmidrule(lr){5-6} \cmidrule(lr){7-8} \cmidrule(lr){9-10}
& &avg@8|64 ($\uparrow$) & Tokens ($\downarrow$) & avg@8|64 ($\uparrow$) & Tokens (\% $\downarrow$) & avg@8|64 ($\uparrow$) & Tokens (\% $\downarrow$) & avg@8|64 ($\uparrow$) & Tokens (\% $\downarrow$) \\
\midrule
\midrule
\multirow{5}{*}{DS-Qwen-2.5-1.5B} 
& AIME24 & 30.10 & 782.3k & 32.50 & 325.9k~(-58.34\%) &\underline{37.50} & \underline{318.2k}~(-59.33\%) & \textbf{37.92} & \textbf{204.3k}~(-73.88\%)\\
& AIME25 & 22.76 & 784.8k & \underline{24.17} & 325.0k~(-58.59\%) & 24.16 & \underline{323.2k}~(-58.82\%) & \textbf{26.67} & \textbf{206.6k}~(-73.68\%)\\
& BRUMO25 & 30.99 & 774.6k & \underline{31.67} & 325.6k~(-57.96\%) & \underline{32.50} & \underline{320.5k}~(-58.62\%) & \textbf{33.75} & \textbf{204.4k}~(-73.61\%) \\
& HMMT25 & 15.05 & 856.4k & 15.00 & 337.2k~(-60.63\%) &\underline{16.67} & \underline{333.8k}~(-61.03\%) & \textbf{17.92} & \textbf{215.5k}~(-74.84\%)\\
& GPQA-D & 33.08 & 550.9k & - & - & - & - & \textbf{48.42} & \textbf{179.4k}~(-67.43\%) \\
\midrule
\multirow{5}{*}{DS-Qwen-2.5-7B} 
& AIME24 & 54.69 & 666.2k & 54.58 & 312.5k~(-53.09\%) & \underline{59.17} & \underline{308.6k}~(-53.68\%) & \textbf{61.67} & \textbf{189.0k}~(-71.63\%) \\
& AIME25 & 39.67 & 703.0k & 39.17 & 317.6k~(-54.82\%) & \underline{37.08} & \underline{315.5k}~(-55.13\%) & \textbf{42.50} & \textbf{197.5k}~(-71.91\%) \\
& BRUMO25 & 50.99 & 656.6k & 51.25 & 312.1k~(-52.46\%) & \underline{53.33} & \underline{309.1k}~(-52.92\%) & \textbf{56.67} & \textbf{190.2k}~(-71.03\%)\\
& HMMT25 & 23.91 & 808.9k & 23.33 & 330.8k~(-59.11\%) & \underline{24.17} & \underline{328.8k}~(-59.35\%) & \textbf{27.08} & \textbf{211.6k}~(-73.84\%)\\
& GPQA-D & 45.95 & 443.8k & - & - & - & - & \textbf{55.75} & \textbf{165.9k}~(-62.61\%)\\
\midrule
\multirow{5}{*}{DS-Qwen-3-8B} 
& AIME24 & 76.93 & 1361k & \underline{78.75} & 398.4k~(-70.73\%) & 77.92 & \underline{396.5k}~(-70.87\%) & \textbf{79.17} & \textbf{279.0k}~(-79.51\%)\\
& AIME25 & 70.68 & 1427k & \underline{72.50} & 408.4k~(-71.39\%) & \textbf{73.33} & \underline{407.5k}~(-71.44\%) & \underline{72.92} & \textbf{290.9k}~(-79.62\%)\\
& BRUMO25 & 75.00 & 1320k & \underline{75.83} & 394.9k~(-70.10\%) & 75.00 & \underline{396.1k}~(-70.01\%) & \textbf{78.75} & \textbf{277.5k}~(-78.98\%)\\
& HMMT25 & 51.04 & 1601k & 50.83 & 427.8k~(-73.28\%) & \underline{52.08} & \underline{427.7k}~(-73.28\%) & \textbf{54.58} & \textbf{311.7k}~(-80.53\%)\\
& GPQA-D & 56.87 & 652.6k & - & - & - & - & \textbf{63.32} & \textbf{193.5k}~(-70.35\%)\\
\midrule
\multirow{5}{*}{GPT-OSS-20B}
& AIME24 & 75.26 & 594.2k & \underline{76.25} & 299.8k~(-49.55\%) & 74.16 & \underline{302.5k}~(-49.09\%) & \textbf{77.50} & \textbf{184.4k}~(-68.98\%) \\
& AIME25 & \underline{70.99} & 673.4k & 69.17 & 311.7k~(-53.71\%) & 69.58 & \underline{310.4k}~(-53.91\%) & \textbf{75.42} & \textbf{191.1k}~(-71.62\%)\\
& BRUMO25 & \underline{68.02} & 575.6k & 66.25 & 298.8k~(-48.09\%) & 67.50 & \underline{297.9k}~(-48.24\%) & \textbf{70.00} & \textbf{183.6k}~(-68.11\%) \\
& HMMT25 & \underline{48.13} & 910.8k & 45.42 & 336.9k~(-63.01\%) & 48.75 & \underline{333.3k}~(-63.41\%) & \textbf{52.92} & \textbf{216.1k}~(-76.27\%)\\
& GPQA-D & 65.55 & 277.2k & - & - & - & - & \textbf{77.46} & \underline{143.4k}~(-48.26\%)\\
\bottomrule
\end{tabular}
}
\end{table*}

\subsection{Detailed Analysis}
Table~\ref{tab:main_results_LaBor} reports results across models and benchmarks. We observe that \textbf{Type~\ref{type:aux}-retrain} consistently outperforms the standard Type~\ref{type:aux} baseline, which is typically trained on public PRM datasets or heuristic labels. This result confirms that MC-estimated soft labels provide a stronger and more informative supervision signal than conventional binary labels, even for external reward models. More importantly, despite being trained on identical data, \textbf{STOP} consistently outperforms \textbf{Type~\ref{type:aux}-retrain} across different model scales. For example, at the 1.5B scale, \textbf{STOP} achieves higher avg@8 on AIME 25 (26.67\% vs.\ 24.16\%) and BRUMO 25 (33.75\% vs.\ 32.50\%), while at the 7B scale it surpasses Type~\ref{type:aux}${}^{\text{retrain}}$ on AIME 24 (61.67\% vs.\ 59.17\%). In addition, while Type~\ref{type:aux} is restricted to mathematical tasks due to its specialized backbone, \textbf{STOP}, implemented via LoRA, naturally generalizes to the scientific domain on GPQA during training, demonstrating greater flexibility. The only exception is a minor difference on DS-Qwen-3-8B for AIME 25 (72.92\% vs.\ 73.33\%), which lies within normal variance; in all other settings, \textbf{STOP} shows clear and consistent advantages.

\subsection{Discussion: The Advantage of Internal Signals}
The superiority of \textbf{STOP} (\textbf{Type~\ref{type:module}}) can be attributed to its ability to mitigate the \emph{information bottleneck} inherent in external evaluation. An external PRM (\textbf{Type~\ref{type:aux}}) judges reasoning quality solely from generated text, which is a discrete and low-dimensional projection of the model’s internal reasoning process and often discards subtle signals of uncertainty and coherence. In contrast, \textbf{STOP} is integrated directly into the generator and has access to dense internal representations, including hidden states and attention patterns. These internal signals preserve rich information about confidence and logical consistency that is largely lost during decoding. By leveraging such first-person internal signals, \textbf{STOP} evaluates the potential of a prefix more accurately than a third-person external reward model.

\section{Derivation and Validation of the Scaling Law}
\label{app:scaling_law}

In Section~\ref{subsec:Power-law}, we introduced the Interaction Scaling Law to describe the relationship among the optimal pruning ratio $\gamma$, the compute budget $C$, and task complexity. In this appendix, we first examine the empirical optimization surfaces that validate this formulation (Appendix~\ref{app:scaling_derivation}), and then provide detailed reference tables for practical deployment (Appendix~\ref{app:gamma_guidelines}).

\subsection{Empirical Observations on Optimal Retention}
\label{app:scaling_derivation}

\begin{figure*}[t]
\vspace{-6mm}
  \centering

  \begin{subfigure}[b]{0.45\linewidth}
    \centering
    \includegraphics[height=4.2cm]{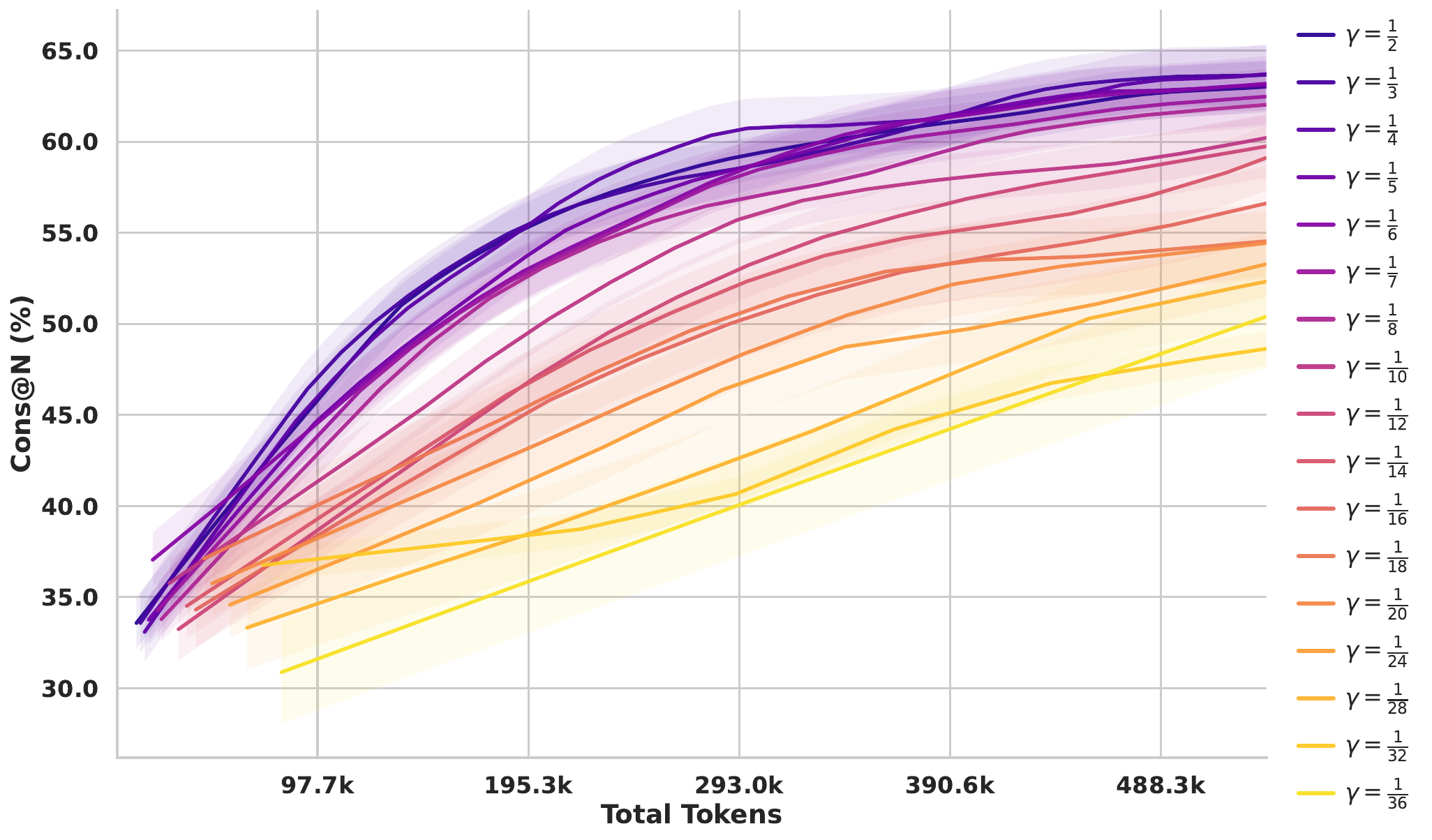}
    \caption{\textbf{AIME 2024 ($L_{prefix}=2048$)}. Optimal $\gamma$ shifts to aggressive pruning as budget increases.}
  \end{subfigure}
  \hfill
  \begin{subfigure}[b]{0.45\linewidth}
    \centering
    \includegraphics[height=4.2cm]{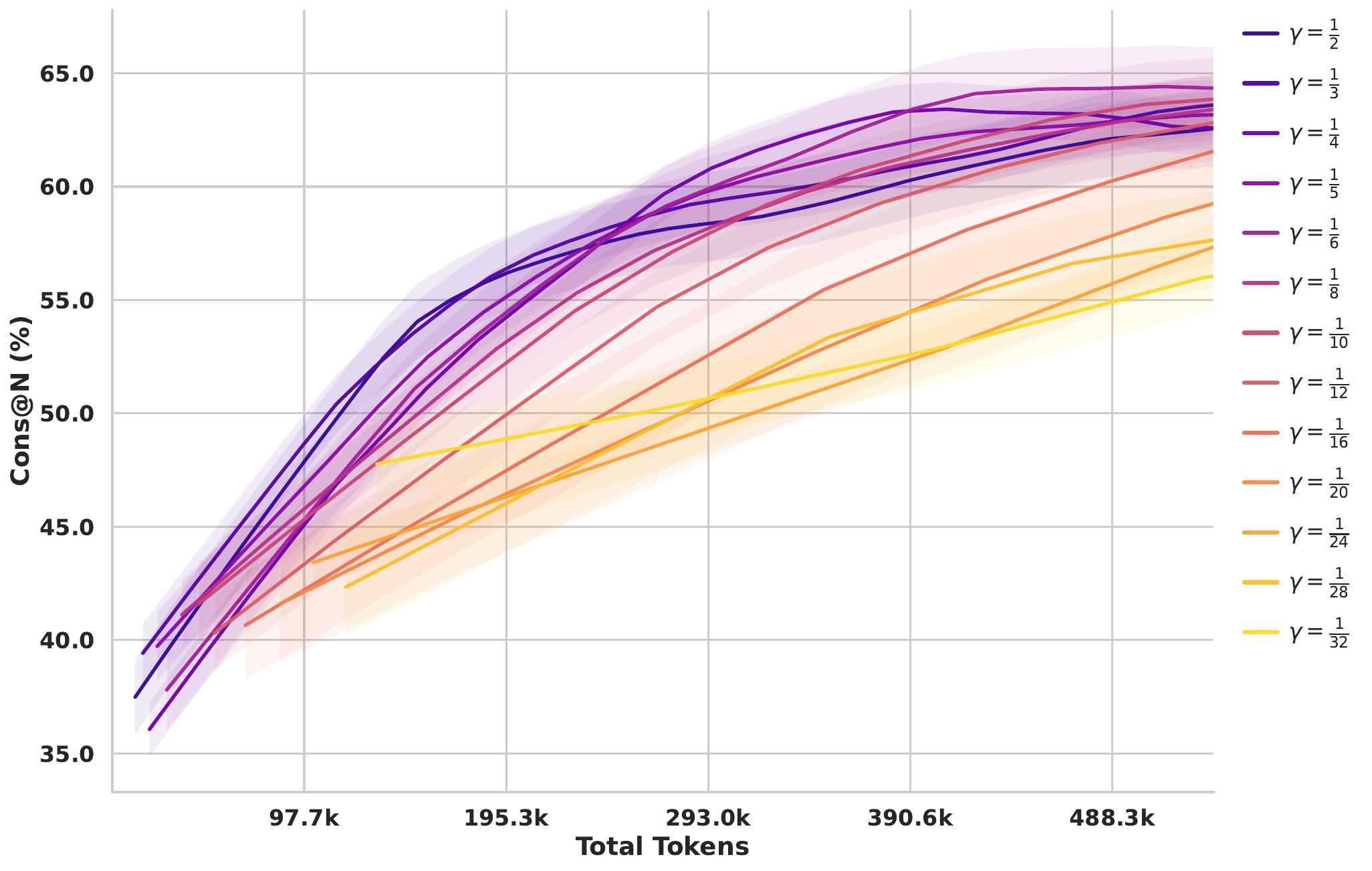}
    \caption{\textbf{AIME 2024 ($L_{prefix}=4096$)}. Longer context enables stable pruning at higher selectivity.}
  \end{subfigure}

  \vspace{0.6em}

  \begin{subfigure}[b]{0.45\linewidth}
    \centering
    \includegraphics[height=4.2cm]{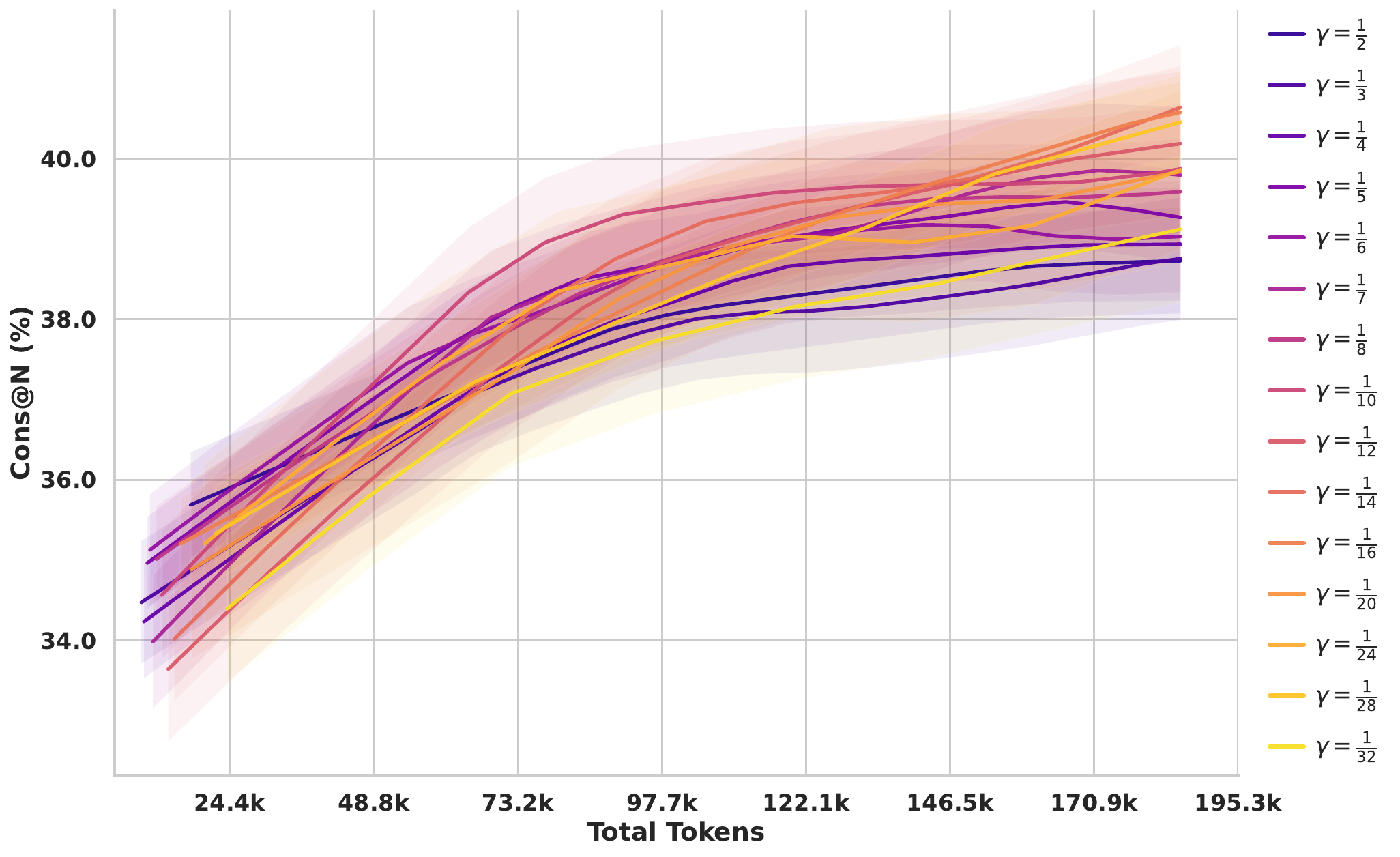}
    \caption{\textbf{GPQA ($L_{prefix}=512$)}. Higher compute budgets drive more aggressive pruning.}
  \end{subfigure}
  \hfill
  \begin{subfigure}[b]{0.45\linewidth}
    \centering
    \includegraphics[height=4.2cm]{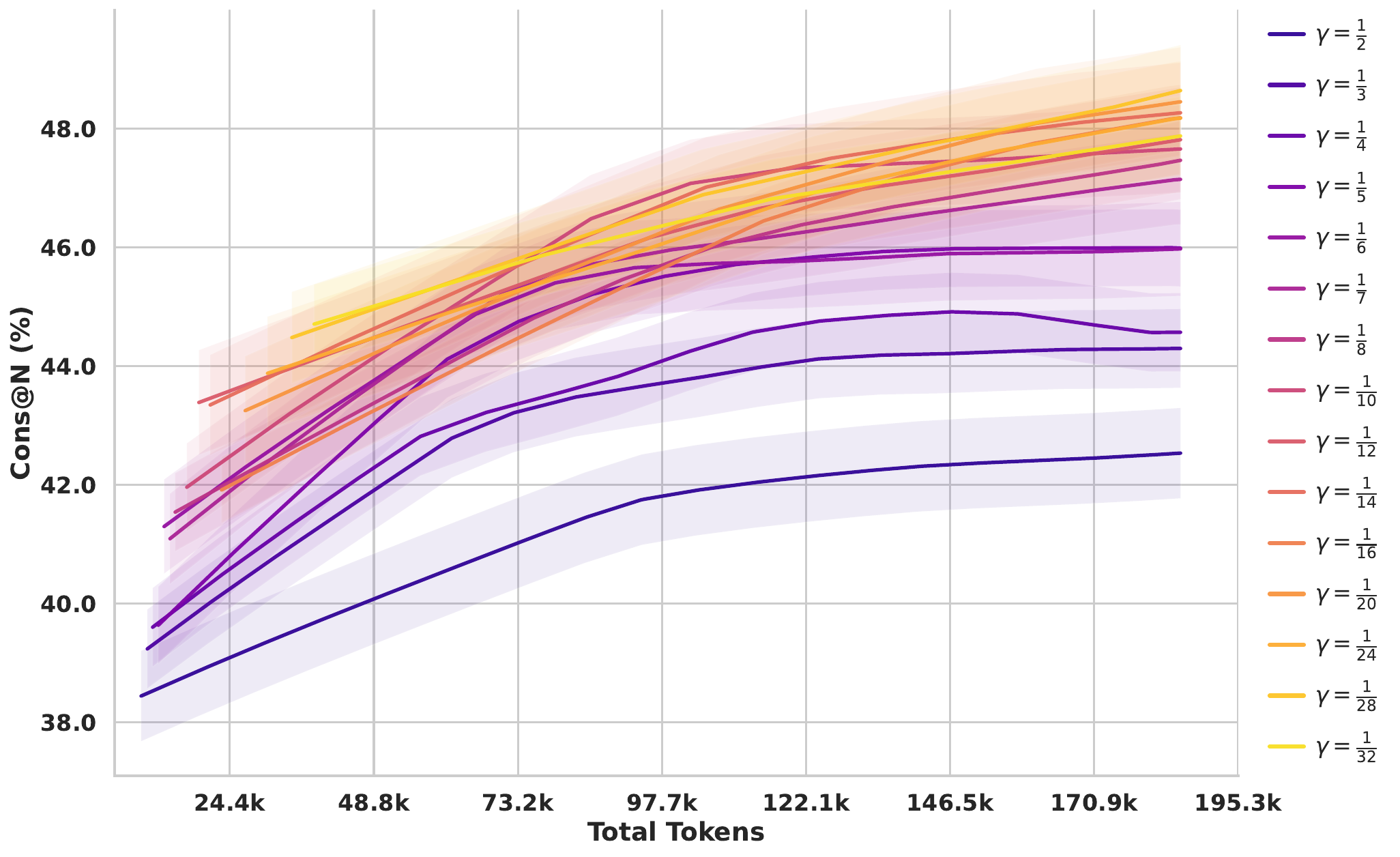}
    \caption{\textbf{GPQA ($L_{prefix}=1024$)}. Scaling behavior remains consistent with longer contexts.}
  \end{subfigure}

  \caption{\textbf{Empirical optimization surfaces.} Impact of retention ratio $\gamma$ across increasing compute budgets.}
  \label{fig:scaling_raw_data}
\end{figure*}

We study how the optimal retention ratio $\gamma^*$, defined as the peak of the performance envelope under a fixed compute budget, varies across benchmarks and prefix lengths $L_{\text{prefix}}$. Visualizations of these empirical surfaces are presented in Figure~\ref{fig:scaling_raw_data}.

\noindent\textbf{Scientific Reasoning (GPQA).}
For GPQA with $L_{\text{prefix}}=512$ and $1024$, the optimal strategy shifts toward more aggressive pruning as the compute budget increases. With short contexts ($L_{\text{prefix}}=512$), $\gamma^*$ is around $1/8$ at low budgets ($\sim 24$k tokens), reflecting a balance between exploration and exploitation. As the budget increases to $195$k tokens, the performance peak moves to smaller values ($\gamma \approx 1/16$), indicating that \textbf{STOP} effectively discards low-quality candidates when sufficient samples are available. For medium contexts ($L_{\text{prefix}}=1024$), conservative retention ($\gamma=1/2$) consistently underperforms. The optimal $\gamma^*$ starts near $1/8$ and rapidly decreases toward $\gamma \approx 1/28$ as compute increases.

This pruning pattern arises from the concise reasoning structure of GPQA. GPQA solutions typically require few steps, so the fixed prefix captures a large portion of the full reasoning trajectory. As a result, the prefix contains high information density and provides a strong pruning signal, enabling \textbf{STOP} to aggressively filter candidates with low risk of removing correct solutions.

\noindent\textbf{Mathematical Reasoning (AIME).}
In contrast, AIME shows a strong dependence on prefix length, reflecting the higher sunk cost of long mathematical derivations. For $L_{\text{prefix}}=2048$, increasing the compute budget shifts the optimal $\gamma^*$ from conservative values ($\gamma \approx 1/2$) toward more aggressive pruning ($\gamma \approx 1/4$). Compared with GPQA, AIME consistently requires higher retention because mathematical reasoning is deeply sequential, and a fixed prefix represents only an initial portion of the full solution, leading to greater downstream uncertainty.

When the context length increases to $L_{\text{prefix}}=4096$, we observe a further shift toward selectivity. Contrary to the expectation that longer contexts require conservative retention, the optimal $\gamma^*$ decreases to the range $\gamma \in [1/6, 1/8]$. This behavior indicates that a longer prefix provides richer evidence for evaluating trajectory quality. With more reasoning history available, the \textbf{STOP} module identifies flawed paths with higher confidence, allowing more aggressive pruning than in the $L_{\text{prefix}}=2048$ setting without sacrificing correct solutions.

\noindent\textbf{Alignment with the Unified Formula.}
These results support the coupled structure of the Interaction Scaling Law. Across all tasks, $\gamma^*$ consistently decreases as the compute budget $C$ increases. At the same time, the optimal pruning level is modulated by the interaction between task domain and available context. Overall, the scaling law adapts to differences in reasoning density across domains and prefix lengths, and it aligns well with the observed empirical optimization landscapes.

\begin{table*}[t]
    \centering
    \small
    \setlength{\tabcolsep}{5pt}
    \caption{\textbf{GPQA (Science, Short-Horizon).} Recommended inverse retention ratios ($\gamma^{-1}$) for tasks with shorter reference lengths ($L_{\text{task}} \approx 8{,}650$). Pruning is more aggressive (higher values) even at lower budgets.}
    \label{tab:gamma_lookup_gpqa}
    \begin{tabular}{c ccccccccc}
    \toprule
    \multirow{2}{*}{\textbf{Prefix Length}} & \multicolumn{9}{c}{\textbf{Compute Budget $C$ (Total Tokens)}} \\
    \cmidrule(lr){2-10}
    ($L_{\text{prefix}}$) & \textbf{140k} & \textbf{160k} & \textbf{180k} & \textbf{200k} & \textbf{220k} & \textbf{240k} & \textbf{260k} & \textbf{280k} & \textbf{300k} \\
    \midrule
    512   & 5.23 & 5.56 & 5.87 & 6.16 & 6.44 & 6.70 & 6.95 & 7.19 & 7.42 \\
    1024  & 6.90 & 7.34 & 7.75 & 8.13 & 8.49 & 8.84 & 9.17 & 9.49 & 9.80 \\
    1536  & 8.11 & 8.63 & 9.11 & 9.56 & 9.99 & 10.40 & 10.79 & 11.16 & 11.52 \\
    2048  & 9.10 & 9.68 & 10.22 & 10.73 & 11.21 & 11.67 & 12.10 & 12.52 & 12.93 \\
    2560  & 9.95 & 10.59 & 11.17 & 11.73 & 12.26 & 12.76 & 13.23 & 13.69 & 14.13 \\
    \bottomrule
    \end{tabular}
\end{table*}

\begin{table*}[t]
    \centering
    \small
    \setlength{\tabcolsep}{5pt}
    \caption{\textbf{AIME (Math, Long-Horizon).} Recommended inverse retention ratios ($\gamma^{-1}$) for tasks with longer reference lengths ($L_{\text{task}} \approx 11{,}950$). Pruning is more conservative (lower values) due to higher reasoning complexity.}
    \label{tab:gamma_lookup_aime}
    \begin{tabular}{c ccccccccc}
    \toprule
    \multirow{2}{*}{\textbf{Prefix Length}} & \multicolumn{9}{c}{\textbf{Compute Budget $C$ (Total Tokens)}} \\
    \cmidrule(lr){2-10}
    ($L_{\text{prefix}}$) & \textbf{200k} & \textbf{250k} & \textbf{300k} & \textbf{350k} & \textbf{400k} & \textbf{450k} & \textbf{500k} & \textbf{550k} & \textbf{600k} \\
    \midrule
    1024  & 1.87 & 2.07 & 2.25 & 2.42 & 2.57 & 2.71 & 2.85 & 2.98 & 3.10 \\
    2048  & 2.47 & 2.73 & 2.97 & 3.19 & 3.39 & 3.58 & 3.76 & 3.93 & 4.09 \\
    3072  & 2.90 & 3.21 & 3.49 & 3.75 & 3.99 & 4.21 & 4.42 & 4.62 & 4.81 \\
    4096  & 3.25 & 3.60 & 3.92 & 4.21 & 4.48 & 4.72 & 4.96 & 5.18 & 5.39 \\
    5120  & 3.56 & 3.94 & 4.29 & 4.60 & 4.89 & 5.17 & 5.42 & 5.66 & 5.90 \\
    \bottomrule
    \end{tabular}
\end{table*}

\subsection{Recommended Retention Guidelines}
\label{app:gamma_guidelines}

Based on the derived scaling law, we provide reference tables for selecting optimal pruning strategies. To \textbf{improve visual clarity and facilitate quick lookup}, we present the guidelines in two separate tables, each corresponding to a different compute budget regime.

These tables are intended primarily as \textbf{illustrative references} for representative task lengths. For other tasks, whether they are similar to GPQA or Math and have different response characteristics, practitioners can directly substitute the task length ($L_{task}$), prefix length ($L_{prefix}$), and compute budget ($C$) into the derived formula (Eq.~\ref{eq:optimial_gamma}) to obtain the exact optimal retention ratio.

Tables~\ref{tab:gamma_lookup_gpqa} and~\ref{tab:gamma_lookup_aime} report the recommended \textbf{inverse retention ratio} ($\gamma^{-1}$) for representative short-horizon tasks ($L_{task} \approx 8{,}650$) and long-horizon tasks ($L_{task} \approx 11{,}950$), respectively.

\section{Detailed Latency and Throughput Benchmarking}
\label{app:latency_details}

In this appendix, we present a detailed analysis of the system efficiency discussed in Section~\ref{subsec:ablation}.
We conduct controlled micro-benchmarks on a single NVIDIA H100 GPU using \textbf{DS-Qwen-2.5-7B}.
The evaluation uses a batch size of 16 and a fixed prefix length of 2,048 tokens to simulate realistic inference conditions.

\subsection{Metric Definitions}
We adopt the following metrics to evaluate computational overhead:
\begin{itemize}[leftmargin=*, noitemsep, topsep=0pt]
    \item \textbf{Generation Time ($T_{\text{gen}}$):} The wall-clock time required for autoregressive decoding of reasoning tokens, excluding any verification operations.
    \item \textbf{Verification Latency ($T_{\text{verify}}$):} The explicit computation time required by the pruning signal generator to produce scores for a batch.
    \item \textbf{System Throughput:} The effective inference speed measured in tokens per second (tok/s). Unlike latency metrics, throughput captures implicit system-level overheads, including CPU--GPU synchronization and pipeline inefficiencies caused by context switching.
\end{itemize}

\subsection{Quantitative Analysis}
Table~\ref{tab:detailed_latency} reports the detailed timing breakdown across different pruning paradigms.
The results reveal a clear mismatch between explicit verification latency and the realized system throughput, especially for heuristic-based methods.

\begin{table*}[t!]
\centering
\small
\caption{\textbf{Breakdown of Inference Latency and Throughput.} 
Note the discrepancy between \textit{explicit cost} and \textit{system impact} for heuristic methods. 
Although Type~\ref{type:heuristic} (SlimSC) shows a low explicit verification cost (1.74\%), the pipeline fragmentation significantly slows down generation, causing a massive \textbf{17.71\% drop in throughput}. 
In contrast, \textsc{STOP} operates in-situ, keeping the throughput drop minimal ($<3\%$) with negligible verification cost (0.59\%).}
\label{tab:detailed_latency}
\resizebox{0.95\linewidth}{!}{
\begin{tabular}{l c c c c c c}
\toprule
\textbf{Method} & \textbf{Gen. Time (s)} & \textbf{Verify Latency (s)} & \textbf{Total Time (s)} & \textbf{Throughput (tok/s)} & \textbf{Throughput Drop ($\downarrow$)} & \textbf{Explicit Verify Cost} \\
\midrule
\textbf{Baseline} (No Pruning) & 33.20 & -- & 33.20 & 986.9 & -- & -- \\
\midrule
Type~\ref{type:heuristic} (SlimSC) & 40.64 & 0.38 & 41.02 & 812.1 & \textbf{17.71\%} & 1.74\% \\
Type~\ref{type:aux} (LaBoR) & 33.53 & 1.13 & 34.68 & 977.3 & 0.97\% & 3.37\% \\
Type~\ref{type:module} (\textbf{STOP}) & 34.13 & \textbf{0.20} & 34.33 & \textbf{960.1} & \textbf{2.71\%} & \textbf{0.59\%} \\
\bottomrule
\end{tabular}
}
\end{table*}

\noindent\textbf{Throughput degradation in heuristic methods.}
A key observation is the pronounced throughput drop in Type~\ref{type:heuristic} (SlimSC). 
Although the cumulative verification latency is small, the method requires frequent similarity computations during chunk-wise generation. 
These repeated interventions fragment GPU kernel execution, prevent sustained high utilization, and increase the base generation time from 33.20s to 40.64s.

\noindent\textbf{Efficiency and implementation of \textsc{STOP}.}
In contrast, the proposed \textsc{STOP} module introduces a minimal verification latency of 0.20s. 
By reusing the resident KV cache, \textsc{STOP} performs verification by processing the sequence $T_s$ in a single forward pass. 
During standard generation, the LoRA adapter remains disabled to strictly preserve the behavior of the base model and is activated only during the verification step. 
The prefix KV cache serves as a shared and immutable reference, and verification appends $T_s$ to a temporary view of this cache to compute the score. 
Once scoring is complete, the temporary branch is discarded. 
This design removes the need for context rollbacks or cache cleanup operations, ensuring that verification introduces no structural overhead into the generation pipeline. 
As a result, the total wall-clock time of \textsc{STOP} (34.33s) remains close to that of the baseline.

\noindent\textbf{Memory Footprint and Deployment Complexity.}
Beyond temporal latency, the spatial overhead of model deployment is a decisive factor.
Methods relying on external verifiers (Type~\ref{type:aux}) impose a \textbf{dual-model burden}: deploying Type~\ref{type:aux} (External PRM) requires hosting a separate PRM alongside the generator.
For example, using a 7B generator with a 7B reward model effectively doubles the VRAM requirement and increases orchestration complexity.
In contrast, \textsc{STOP} is implemented as a lightweight LoRA adapter attached directly to the frozen generator.
This \textbf{integrated architecture} adds only a minimal number of parameters, incurring \textbf{negligible additional VRAM overhead} for model weights.
It eliminates the need for managing secondary inference services, making \textsc{STOP} a "plug-and-play" solution for existing pipelines.

\section{Extended Attention Analysis}
\label{app:mechanistic_interpretability}

\begin{figure*}[h]
  \centering
  \begin{subfigure}[b]{0.48\linewidth}
    \centering
    \includegraphics[height=5.0cm, keepaspectratio]{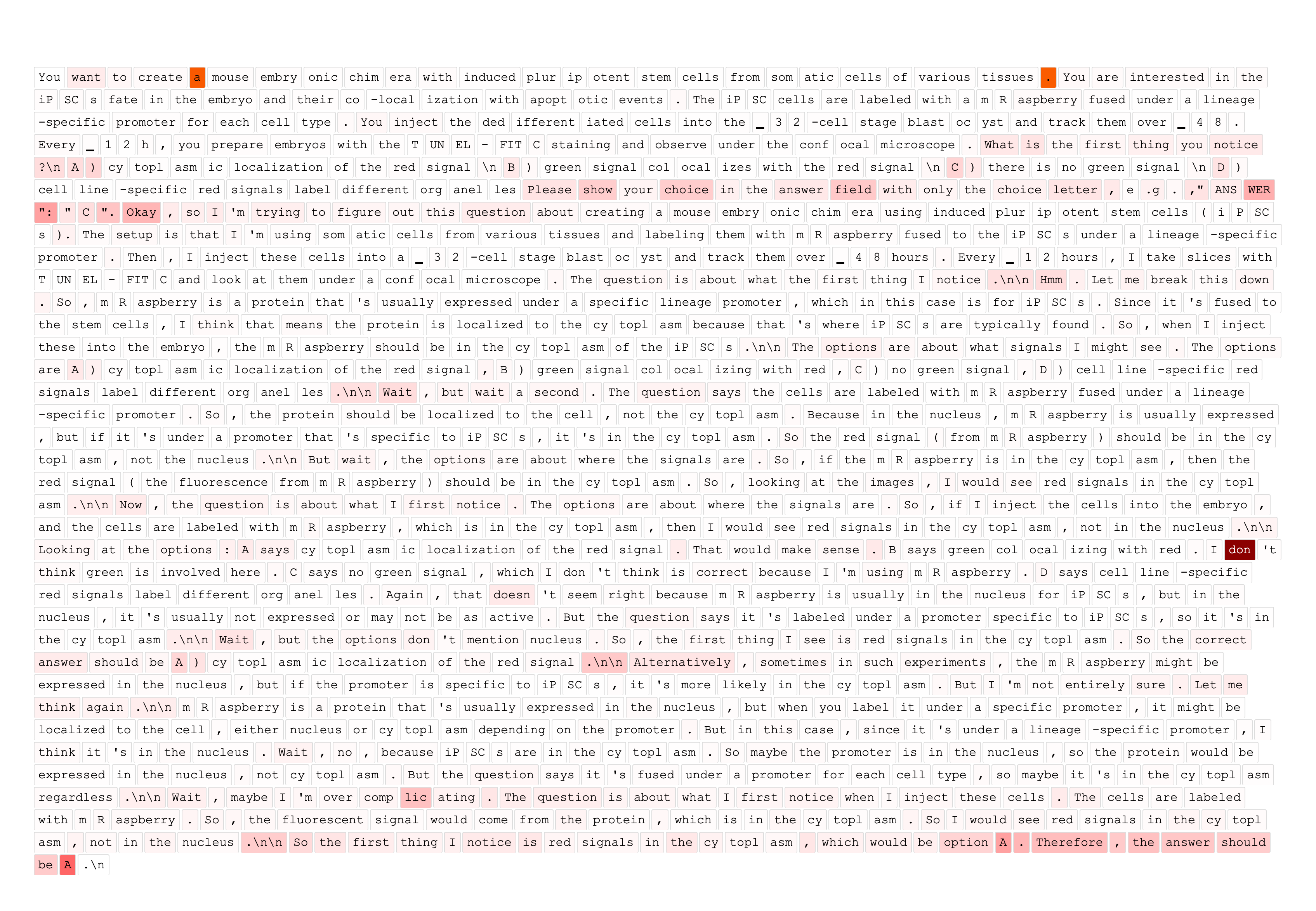}
    \caption{\textbf{High-scoring Case.} The module focuses on the logical negation ``don't'' (a cognitive pivot) rather than simply jumping to the answer option.}
    \label{fig:viz_high_1}
  \end{subfigure}
  \hfill
  \begin{subfigure}[b]{0.48\linewidth}
    \centering
    \includegraphics[height=5.0cm, keepaspectratio]{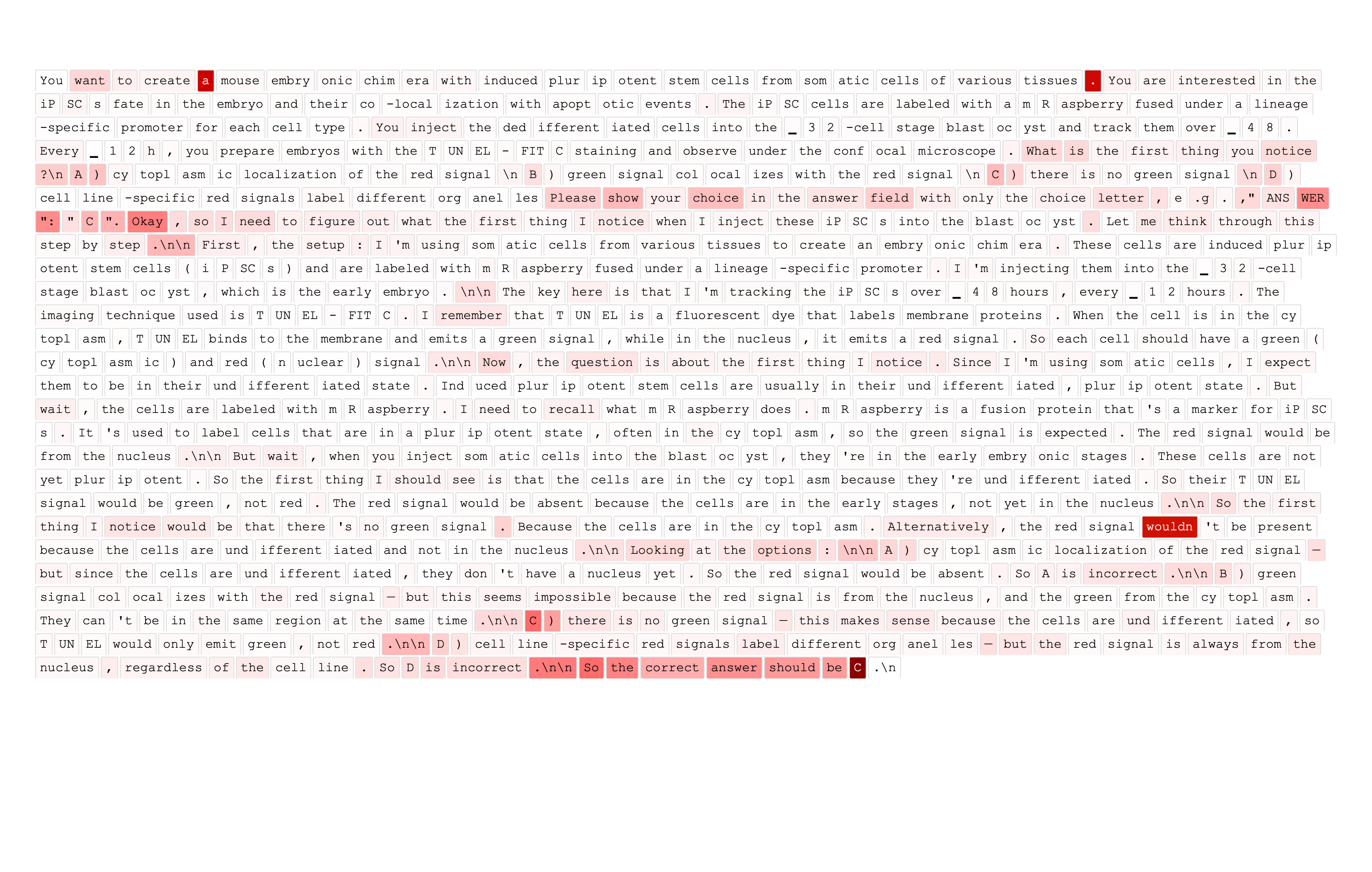}
    \caption{\textbf{Low-scoring Case.} Attention concentrates heavily on the answer option itself (``C''), ignoring the sparse reasoning context.}
    \label{fig:viz_low_1}
  \end{subfigure}

  \vspace{1em}

  \begin{subfigure}[b]{0.48\linewidth}
    \centering
    \includegraphics[height=5.0cm, keepaspectratio]{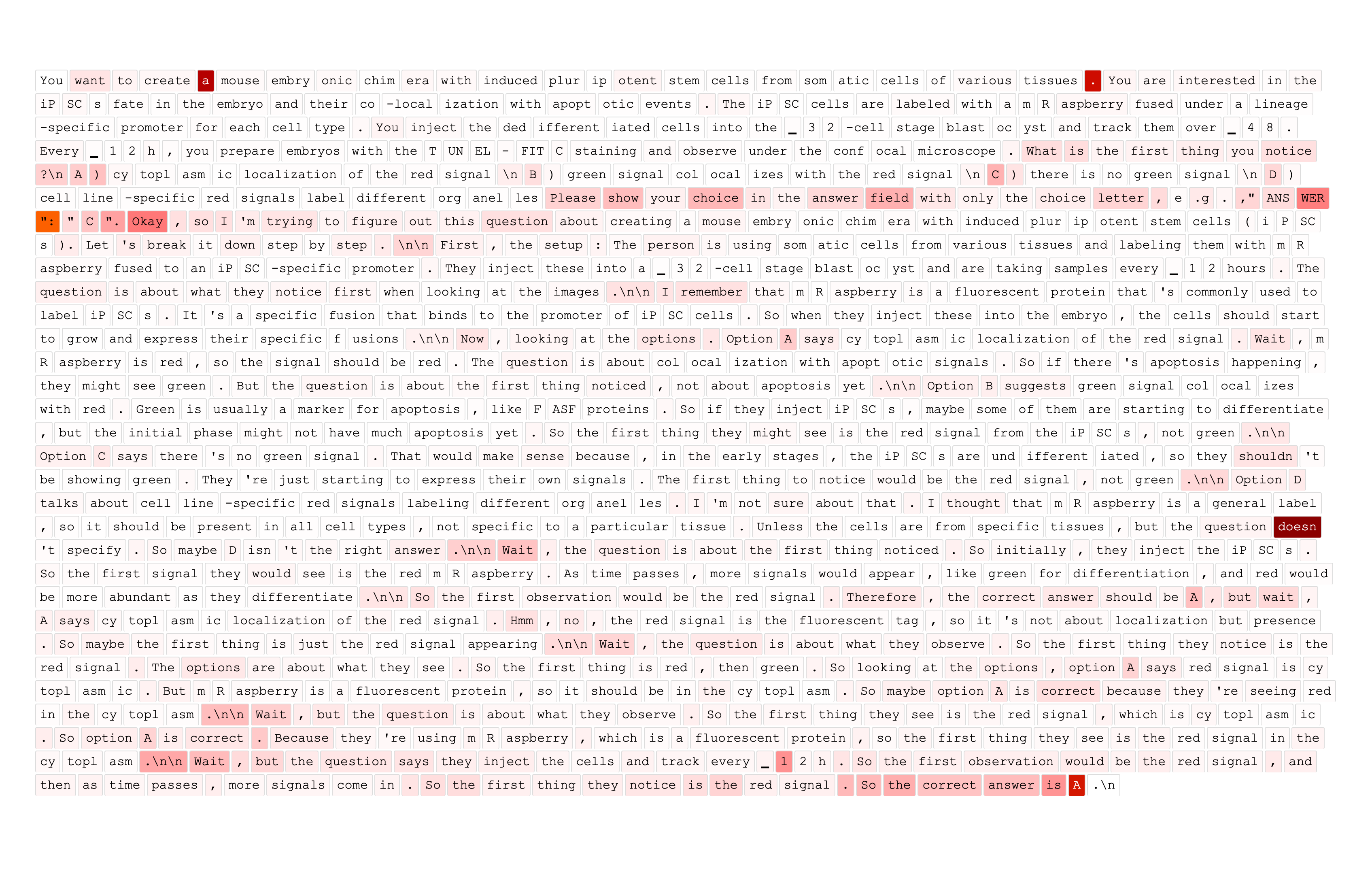}
    \caption{\textbf{High-scoring Case.} Similar to (a), the module attends to the logical marker ``doesn't,'' prioritizing the validity of the reasoning process over the final outcome.}
    \label{fig:viz_high_2}
  \end{subfigure}
  \hfill
  \begin{subfigure}[b]{0.48\linewidth}
    \centering
    \includegraphics[height=5.0cm, keepaspectratio]{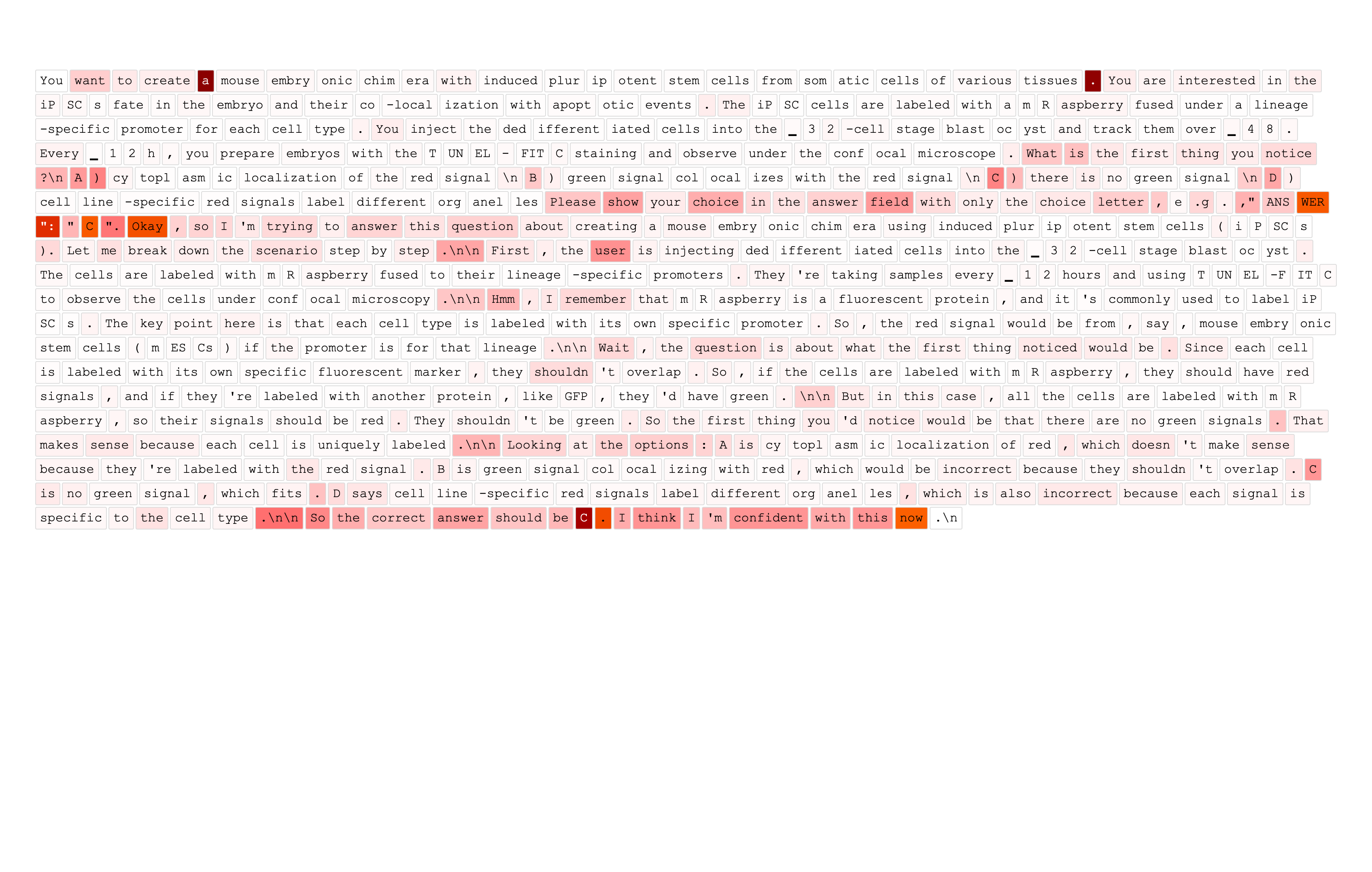}
    \caption{\textbf{Low-scoring Case.} The module demonstrates premature closure by fixating on the terminal choice (``C'') while bypassing critical logical intermediates.}
    \label{fig:viz_low_2}
  \end{subfigure}

  \caption{
  \textbf{Extended Visualization of \texttt{[STOP]} Attention Maps.}
  While \textbf{STOP} broadly tracks structural markers (e.g., ``Wait'', ``Therefore'') in all cases, it distinguishes reasoning quality by focus: \textbf{High-scoring paths} (left) prioritize logical pivots (e.g., ``don't''), whereas \textbf{Low-scoring paths} (right) exhibit \textbf{premature closure} by fixating on the terminal answer options.
  }
  \label{fig:more_cases}
\end{figure*}

In Section~\ref{subsec:interpretability}, we hypothesize that the \textbf{STOP} module acts as a process-oriented evaluator. To empirically validate this, we analyze the attention patterns in Figure~\ref{fig:more_cases}.

\noindent\textbf{Universal Attention Pattern.}
Consistent with the findings in Section 5.3, \textbf{STOP} exhibits a broad attention pattern across all samples. Regardless of the score, the module consistently tracks structural discourse markers (e.g., ``Wait'', ``Hmm'', ``Therefore'', ``but'', ``\textbackslash n\textbackslash n'') as well as the final answer text. This confirms that the module monitors the structural progression of the reasoning chain.

\noindent\textbf{Distinguishing Quality via Attention Focus.}
However, a critical distinction determines the quality score.
In \textbf{High-Scoring Trajectories} (Figures~\ref{fig:more_cases}\subref{fig:viz_high_1} and \subref{fig:viz_high_2}), attention prioritizes \textbf{logical negations} (e.g., ``don't'' and ``doesn't'')—which serve as cognitive pivots—over the final answer options, indicating that \textbf{STOP} values the validity of the logical derivation.
Conversely, \textbf{Low-Scoring Trajectories} (Figures~\ref{fig:more_cases}\subref{fig:viz_low_1} and \subref{fig:viz_low_2}) exhibit a pattern of \textbf{premature closure}: attention disproportionately fixates on the \textbf{answer options themselves} (e.g., the token ``C'') while neglecting the reasoning context, serving as a robust signal for identifying guessing behavior.

\end{document}